%% file: iclr2027_conference.tex
\pdfoutput=1

\documentclass{article} %
\usepackage{iclr2027_conference,times}

\input{math_commands.tex}

\usepackage{hyperref}
\usepackage{url}
\usepackage{graphicx}
\usepackage{booktabs}
\usepackage{multirow}
\usepackage{makecell}
\usepackage{arydshln}
\usepackage{enumitem}
\usepackage{tcolorbox}
\usepackage{pgfplots}
\usepackage{flafter}
\usepackage{wrapfig}
\usepackage{subcaption}
\usepackage{float}
\usepackage{algorithm}
\usepackage[noend]{algpseudocode}
\algrenewcommand\algorithmicrequire{\textbf{Input:}}
\algrenewcommand\algorithmicensure{\textbf{Output:}}
\pgfplotsset{compat=1.16}

\makeatletter
\renewcommand\makecell[2][c]{\begingroup
  \def\mcell@L{\hfil}\def\mcell@R{\hfil}%
  \if l#1\let\mcell@L\relax\fi \if r#1\let\mcell@R\relax\fi
  \def\\{\cr}%
  \leavevmode\hbox{$\m@th\vcenter{\baselineskip\z@\lineskip\z@
    \halign{\mcell@L\strut##\mcell@R\cr#2\crcr}}$}\endgroup}
\def\@mkpream#1{\let\adl@colhtdp\relax
  \let\adl@vlineL\relax \let\adl@vlineR\relax
  \let\adl@startmbox\relax \let\adl@endmbox\relax
  \global\adl@ncol\@ne \adl@leftruletrue
  \adl@mkpream{#1}\adl@columns\adl@ncol \aftergroup\adl@preaminit}
\makeatother

\definecolor{cvprblue}{rgb}{0.21,0.49,0.74}
\definecolor{priorcolor}{RGB}{178,34,34}
\newcommand{\prior}[1]{\textcolor{priorcolor}{#1}}

\hypersetup{
    colorlinks=true,
    linkcolor=cvprblue,
    citecolor=cvprblue,
    urlcolor=cvprblue
}

\title{
   When to Switch: Reliable Action-Chunk Extension for Vision-Language-Action Models
   }

\author{%
Seonghoon Yu$^{1}$\quad Dongwon Kim$^{1}$\quad HyungRok Jung$^{2}$\quad Yoonjae Baek$^{4}$\\
\textbf{Byung-kwan Lee$^{3}$\quad Suha Kwak$^{4}$\quad Jeany Son$^{4}$}\\[4pt]
{\normalfont $^{1}$KAIST\quad $^{2}$GIST\quad $^{3}$NVIDIA\quad $^{4}$POSTECH}
}

\newcommand{\ie}{\textit{i.e.}}
\newcommand{\eg}{\textit{e.g.}}

\iclrfinalcopy
\begin{document}

\maketitle
\lhead{Preprint}

\input{sec/00_abstract}
\input{sec/01_intro.tex}
\input{sec/02_related.tex}
\input{sec/03_method.tex}

\input{sec/05_experiment.tex}
\input{sec/10_conclusion.tex}

\bibliography{iclr2027_conference}
\bibliographystyle{iclr2027_conference}

\input{sec/XX_appendix.tex}

\end{document}

%% file: math_commands.tex
\usepackage{amsmath,amsfonts,bm}

\def\eqref#1{equation~\ref{#1}}

\def\1{\bm{1}}

\DeclareMathAlphabet{\mathsfit}{\encodingdefault}{\sfdefault}{m}{sl}
\SetMathAlphabet{\mathsfit}{bold}{\encodingdefault}{\sfdefault}{bx}{n}

%% file: sec/00_abstract.tex
\vspace{-0.35cm}

\begin{abstract}

\vspace{-0.15cm}

Vision-Language-Action (VLA) models serve as unified policies for robotic manipulation,
yet their expensive inference forces robots to pause between policy calls,
resulting in stop-and-go execution that interrupts smooth motion and prolongs task completion.
Extending the action chunk reduces policy calls and hence these pauses,
but predicting farther into the future makes long-chunk execution unreliable.
To understand where this unreliability arises,
we analyze action errors within long chunks and find that they concentrate around transitions
between manipulation subskills, growing sharply with chunk length.
This suggests the importance of \emph{transition timing}, \ie, when to switch subskills within a chunk.
Motivated by this observation,
we introduce RACE (\textbf{R}eliable \textbf{A}ction-\textbf{C}hunk \textbf{E}xtension), 
a framework that predicts the transition timing from an auxiliary one-step denoising pass
and conditions action generation on it.
By learning and conditioning on transition timing,
RACE reduces errors at subskill transitions and enables reliable execution of longer action chunks.
Across simulation benchmarks, RACE outperforms fine-tuning at the same chunk length;
with $2\times$ longer chunks, it surpasses recent state-of-the-art and efficient VLAs in success rate,
and with $4\times$ longer chunks, it remains competitive.
On a real robot, RACE uses $4\times$ longer chunks,
which reduces the idle time caused by stop-and-go execution by about $5\times$,
while achieving a higher success rate than fine-tuning with the same chunk length.
Code and a real-robot demo are available on
\href{https://github.com/Seonghoon-Yu/RACE-VLA}{Github}.

\end{abstract}

%% file: sec/01_intro.tex
\vspace{-0.4cm}
\section{Introduction}
\vspace{-0.2cm}

Vision-language-action (VLA) models~\citep{rt2, octo, roboflamingo, openvla, pi0}, 
such as $\pi_{0.5}$~\citep{pi05} and GR00T~\citep{gr00t}, 
serve as unified policies for diverse robotic manipulation tasks by generating actions on top of large-scale vision-language models.
However, their large scale makes each policy inference computationally expensive.
Consequently, the robot often finishes executing its current actions before the next ones are ready,
forcing repeated pauses between policy calls that disrupt continuous robot operation, \ie, stop-and-go execution~\citep{rtc,vlash}.
Over hundreds of control steps, these pauses accumulate into \emph{idle time},
which prolongs task completion and interrupts smooth robot motion,
hindering the real-world deployment of VLAs.

\setlength{\columnsep}{6pt}
\begin{wrapfigure}{r}{0.35\linewidth}
    \vspace{-12.5pt}
    \centering
    \includegraphics[width=\linewidth]{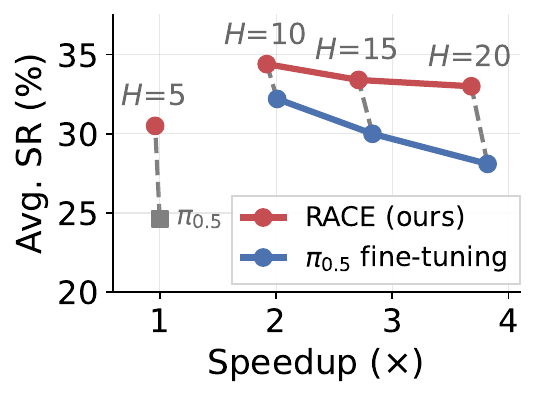}
    \vspace{-24.2pt}
    \caption{Effect of chunk length $H$ on VLABench, with per-episode inference speedup over $\pi_{0.5}$ (Appendix~\ref{app:speedup_measure}).}
    \label{fig:tradeoff}
    \vspace{-3pt}
\end{wrapfigure}
Action chunking~\citep{act,diffusionpolicy,pi0} mitigates these pauses by predicting multiple consecutive actions per policy call,
keeping the robot in motion longer between policy inferences.
Extending this chunk~\citep{policytrim} further reduces the number of policy calls and the time lost to these pauses,
making robot execution more continuous.
Moderately longer chunks even improve task success,
which has been attributed to temporal consistency~\citep{bid} or implicit ensembling~\citep{whychunking}. 
Pushing the chunk further, however, forces the policy to predict farther into the future
and to execute longer without intermediate feedback (\ie, open-loop execution).
Beyond a certain length, this makes execution unreliable and success rates decline~\citep{autohorizon,adaptivechunking} (Fig.~\ref{fig:tradeoff}).

\begin{figure*}[t]
    \centering
    \begin{subfigure}{1.0\textwidth}
        \centering
        \includegraphics[width=\linewidth]{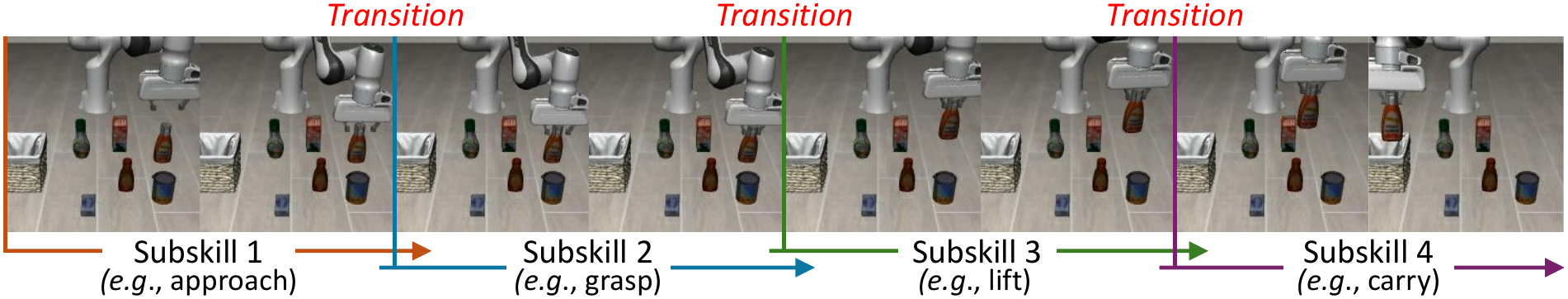}
        \vspace{-0.45cm}
        \caption{Transition between manipulation subskills}
        \vspace{-0.05cm}
        \label{fig:motivation_a}
    \end{subfigure}\\[4pt]
    \begin{subfigure}{0.48\textwidth}
        \centering
        \includegraphics[width=0.9\linewidth]{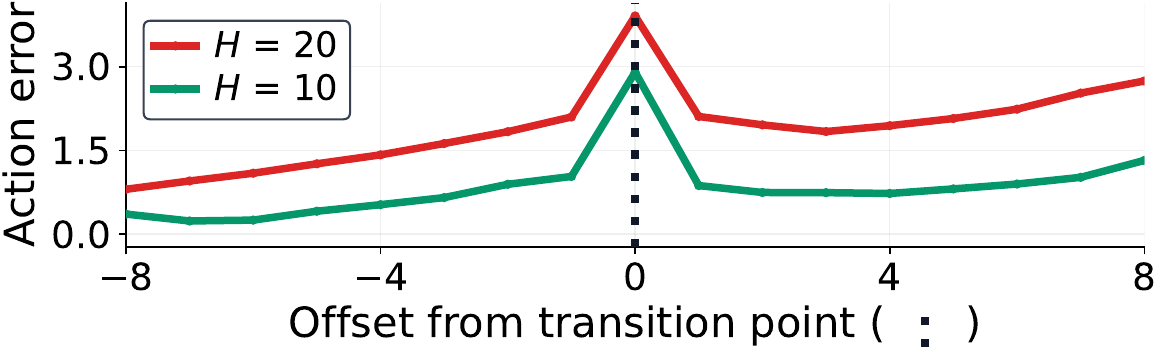}
        \vspace{-0.2cm}
        \caption{Transition-localized errors, $\pi_{0.5}$}
        \label{fig:motivation_b}
    \end{subfigure}
    \hfill
    \begin{subfigure}{0.48\textwidth}
        \centering
        \includegraphics[width=0.9\linewidth]{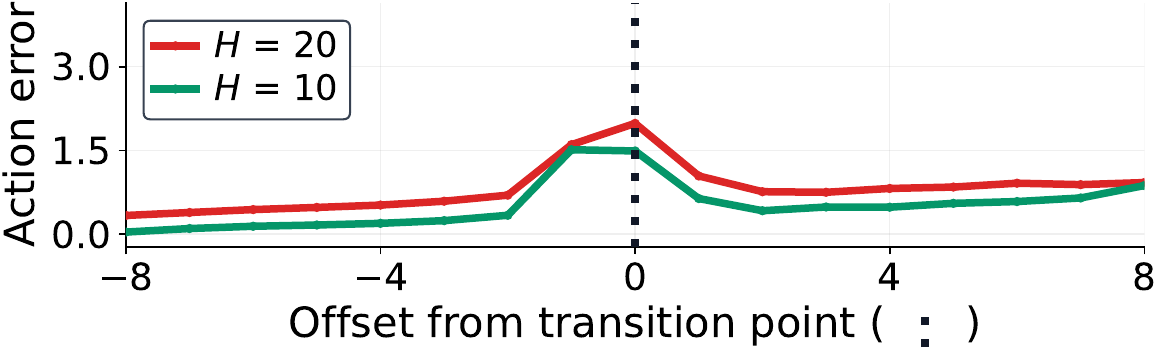}
            \vspace{-0.2cm}
        \caption{Transition-localized errors, RACE (ours)}
        \label{fig:motivation_c}
    \end{subfigure}
    \vspace{-0.2cm}
    \caption{
        Action prediction errors around subskill transitions:
        (a) an example of transitions between manipulation subskills; 
        (b)~action errors spike at the transition point within a chunk,
        and these spikes grow with the chunk length $H$; 
        and
        (c)~RACE reduces these error spikes across chunk lengths $H$.
    }
    \vspace{-0.5cm}
    \label{fig:motivation}
\end{figure*}

To better understand why long-chunk execution becomes unreliable, we examine how action errors are distributed within a chunk. 
Besides the gradual error growth expected from open-loop execution,
we also observe pronounced error spikes around \emph{transition points},
the steps at which the robot switches from one manipulation subskill (\eg, approaching, grasping, or lifting) to the next (Fig.~\ref{fig:motivation_a}).
Since demonstrations do not annotate subskills, 
we detect these points as abrupt changes in action dynamics using PELT~\citep{pelt}, following prior work~\citep{cotpc,primt} (Appendix~\ref{app:transition}).
The error spikes at these transition points grow substantially with chunk length (Fig.~\ref{fig:motivation_b}; measured as in Appendix~\ref{app:action_error_measure}).
As the chunk grows longer, 
the policy must anticipate these subskill switches farther in advance without intermediate feedback, 
making these switches increasingly difficult to predict. 
This suggests that \emph{transition timing}, \ie, when to switch subskills within a chunk,
is an important challenge in long-chunk execution.

Motivated by this observation, we introduce RACE
(\textbf{R}eliable \textbf{A}ction-\textbf{C}hunk \textbf{E}xtension),
a framework that enables VLAs to reliably execute longer action chunks
by learning to predict transition timing and conditioning action generation on it.
Learning to predict transition timing makes the action expert aware of upcoming transitions,
and conditioning on the predicted timing further informs every denoising step of where transitions are expected within the chunk,
together reducing errors around transitions.
To achieve this, RACE proceeds in two stages:
1) \textbf{Transition-timing Prediction via One-step Denoising}, 
which estimates a transition-timing prior from an auxiliary one-step denoising pass and the cached VLM features,
supervised by transition labels automatically extracted from demonstrations;
and 2) \textbf{Transition-conditioned Action Generation}, which injects the prior into every denoising step
through token-wise modulation of the action expert, with learnable gates controlling its strength per step.
Since error spikes appear at transitions at every chunk length (Fig.~\ref{fig:motivation_b}),
RACE reduces these errors across chunk lengths (Fig.~\ref{fig:motivation_c}),
improving task success not only for longer chunks but at each chunk length (Fig.~\ref{fig:tradeoff}).

In our experiments, RACE outperforms fine-tuning at the same chunk length on manipulation benchmarks;
with $2\times$ longer chunks, it achieves higher success rates than recent state-of-the-art VLAs and efficient VLA approaches,
and with $4\times$ longer chunks, it remains competitive with them.
On a real robot, using $4\times$ longer chunks reduces idle time by about $5\times$,
and RACE achieves a higher success rate than fine-tuning with the same chunk length.

Our contributions are summarized as follows:
\vspace{-0.25cm}
\begin{itemize}[leftmargin=10pt]
    \item
    We identify transitions between manipulation subskills 
    as where action errors concentrate when extending action chunks,
    suggesting the importance of \emph{transition timing} within a chunk.

    \item
    We introduce RACE, a framework for \textbf{R}eliable \textbf{A}ction-\textbf{C}hunk \textbf{E}xtension
    that learns to predict transition timing from an auxiliary one-step denoising pass, making the action expert transition-aware,
    and conditions generation on it via token-wise modulation with per-step learnable gates.

    \item
    Experiments show that RACE achieves higher success rates than recent state-of-the-art and efficient VLA approaches
    with $2\times$ longer action chunks and remains competitive with $4\times$ longer chunks,
    and enables longer chunks that reduce idle time on a real robot.

\end{itemize}

%% file: sec/02_related.tex
\vspace{-0.25cm}
\section{Related Work}
\vspace{-0.25cm}

\paragraph{Vision-Language-Action Models.}
VLAs adapt pretrained vision-language models to robot control using diverse robot demonstrations~\citep{univla}.
Early models such as RT-2~\citep{rt2} and OpenVLA~\citep{openvla} generate discrete action tokens autoregressively,
while later models predict continuous action chunks
through parallel decoding~\citep{openvlaoft} or flow-matching action experts~\citep{pi05,gr00t}.
RACE reduces the pauses caused by their inference latency by making longer action chunks reliable.

\vspace{-0.25cm}
\paragraph{Action Chunking and Asynchronous Execution.}
Action chunking~\citep{act,diffusionpolicy} predicts multiple future actions per policy call,
but later actions in a longer chunk become unreliable without new observations.
\textit{Adaptive chunking methods}~\citep{adaptivechunking,autohorizon,dvac} therefore execute only the reliable prefix of each chunk.
PACE~\citep{pace} shortens execution around transitions detected from low-speed valleys, 
whereas RACE makes those actions reliable so that the whole chunk can be executed.
Closest to our work, PolicyTrim~\citep{policytrim} extends the chunk through reinforcement learning at a substantial training cost,
whereas RACE requires only lightweight post-training of the action expert (Appendix~\ref{app:training_efficiency}).
\textit{Asynchronous execution}~\citep{rtc,vlash,remac} hides inference latency by generating the next chunk during execution,
but some idle time remains when inference takes longer than executing a chunk;
longer chunks can further reduce it (Sec.~\ref{subsec:real_world}), so RACE is complementary to asynchronous execution.

\vspace{-0.25cm}
\paragraph{Efficient VLA Inference.}
Another line of work reduces the cost of each policy call by pruning visual tokens~\citep{vlapruner,safepruner,grids} or reusing computation across consecutive calls~\citep{vlacache,latentbridge}.
These methods leave the number of policy calls unchanged, whereas RACE reduces it; the two are orthogonal and can be combined.

\vspace{-0.25cm}
\paragraph{Subskill Structure in Policy Learning.}
Demonstrations are often decomposed into subskills,
through either change-point detection on action trajectories~\citep{pelt,cotpc}
or manual stage annotation~\citep{sage,bihil}.
Prior work injects this structure into the policy
as a prediction target, by predicting the goal state of each subskill~\citep{cotpc};
as an input, by conditioning on the inferred stage or within-subtask progress~\citep{sage,bihil};
or as an auxiliary output, by generating a progress sequence jointly with actions~\citep{palm}.
RACE instead supervises the action expert to predict the transition timing within each chunk,
encouraging transition-aware action representations,
and further conditions action generation on the predicted timing.

\begin{figure}[t]
    \centering
    \includegraphics[width=1.0\textwidth]{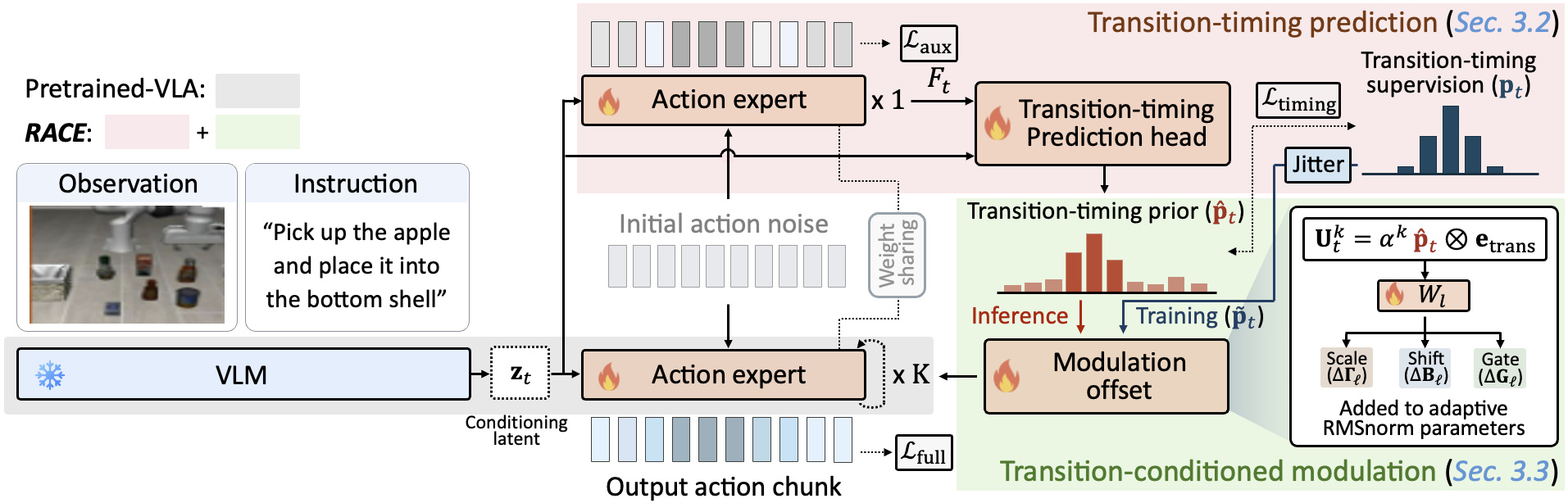}
    \vspace{-0.55cm}
    \caption{Overview of RACE.
    Given an observation and instruction, the frozen VLM extracts features cached for both passes.
    RACE predicts a transition-timing prior via an auxiliary one-step denoising pass (Sec.~\ref{sec:head}),
    then restarts denoising from the same noise, conditioned on it (Sec.~\ref{sec:cond}).
    }
    \vspace{-0.45cm}
    \label{fig:overall_framework}
\end{figure}

%% file: sec/03_method.tex
\vspace{-0.3cm}
\section{RACE: Reliable Action-Chunk Extension}
\vspace{-0.3cm}
\label{sec:method}

In this section, we present RACE, a framework that
enables VLAs to reliably execute longer action chunks (Fig.~\ref{fig:overall_framework}).
We first give an overview of how RACE augments flow-matching action generation with a transition-timing prior (Sec.~\ref{sec:overview}),
and then describe how this prior is predicted from an auxiliary one-step denoising pass (Sec.~\ref{sec:head})
and how it conditions action generation (Sec.~\ref{sec:cond}).

\vspace{-0.3cm}
\subsection{Overview of RACE}
\vspace{-0.25cm}
\label{sec:overview}

Our framework builds upon a flow-matching VLA, such as $\pi_{0.5}$~\citep{pi05},
which consists of a vision-language model (VLM) and an action expert.
At each policy call $t$, the VLM first encodes the current visual observation $\mathbf{o}_t$ and the language instruction $\boldsymbol{\ell}$ into context cache features $\mathbf{z}_t = \text{VLM}(\mathbf{o}_t, \boldsymbol{\ell})$.
Conditioned on $\mathbf{z}_t$, 
the action expert generates an action chunk $\mathbf{A}_t=(\mathbf{a}_t,\ldots,\mathbf{a}_{t+H-1})$ of length $H$
by iteratively transforming a sequence of Gaussian noise $\mathbf{A}_t^{0}\sim \mathcal{N}(\mathbf{0}, \mathbf{I})$ over $K$ denoising steps:
$\mathbf{A}_t^{k} = \mathbf{A}_t^{k-1} + \tfrac{1}{K}\, v_{\theta}(\mathbf{A}_t^{k-1}, \mathbf{z}_t)$
for $k = 1, \ldots, K$, 
where $v_{\theta}$ is the velocity field predicted by the action expert, which also takes the flow time $(k-1)/K$ as input (omitted for brevity),
and $\mathbf{A}_t^{k}$ denotes the intermediate chunk after the $k$-th denoising step,
with the final action chunk $\mathbf{A}_t = \mathbf{A}_t^{K}$.

RACE augments this process with a transition-timing prior $\hat{\mathbf{p}}_t$,
an estimate of where subskill transitions occur within the chunk,
and conditions every denoising step on it, as follows:
\setlength{\abovedisplayskip}{3pt}
\setlength{\belowdisplayskip}{3pt}
\setlength{\abovedisplayshortskip}{0pt}
\setlength{\belowdisplayshortskip}{0pt}
\begin{align}
    \mathbf{A}_t^{k}
    =
    \mathbf{A}_t^{k-1}
    +
    \frac{1}{K}\,
    v_{\theta}\!\left(
        \mathbf{A}_t^{k-1},
        \mathbf{z}_t,
        \alpha^{k}\,\prior{\hat{\mathbf{p}}_t}
    \right),
    \qquad
    k = 1, \ldots, K,
    \label{eq:race_denoising}
\end{align}
where $\alpha^{k} \in (0,1)$ is a learnable per-step gate with sigmoid activation that controls how strongly
the prior is injected at each denoising step.
During training, the flow time is sampled continuously as in $\pi_{0.5}$ rather than at the $K$ inference steps,
so we use the gate $\alpha^{k}$ of the step whose interval $[\tfrac{k-1}{K}, \tfrac{k}{K})$ contains it.
RACE thus runs three steps:
(1) an auxiliary one-step pass from $\mathbf{A}_t^{0}$ without the prior yields action-expert features;
(2) a head estimates $\hat{\mathbf{p}}_t$ from them and $\mathbf{z}_t$ (Sec.~\ref{sec:head});
and (3) the full denoising restarts from the same $\mathbf{A}_t^{0}$, conditioning every step on $\hat{\mathbf{p}}_t$ (Sec.~\ref{sec:cond}).
During training, the jittered target $\tilde{\mathbf{p}}_t$ replaces $\hat{\mathbf{p}}_t$ in step (3),
while the head is supervised with the target $\mathbf{p}_t$ (Sec.~\ref{sec:cond}).
Only the action expert and the added modules are trained with the VLM frozen,
and inference adds only the auxiliary pass and the head to each policy call, both reusing the cached VLM features.

\vspace{-0.15cm}
\subsection{Transition-timing Prediction via One-step Denoising}
\vspace{-0.2cm}
\label{sec:head}

To condition action generation on a transition-timing prior $\hat{\mathbf{p}}_t$,
RACE predicts this prior from two sources of information with a learnable transition-timing prediction head:
(1) \textbf{the cached VLM features $\mathbf{z}_t$}, computed in Sec.~\ref{sec:overview},
which encode contextual information about the current observation and instruction.
(2) \textbf{hidden states from the action expert}, extracted through an auxiliary one-step denoising pass starting from initial noise $\mathbf{A}_t^{0}$.
While the VLM features provide contextual cues,
the auxiliary hidden states can offer preliminary motion information aligned with each action position in the chunk.
We next describe how the head predicts $\hat{\mathbf{p}}_t$ from these two sources.

\vspace{-0.2cm}
\paragraph{Auxiliary One-step Denoising Pass.}
Before the full denoising process of Eq.~(\ref{eq:race_denoising}),
RACE performs an auxiliary one-step denoising pass on the initial noise sample $\mathbf{A}_t^{0}$,
computing the velocity $\mathbf{v}_t^{\mathrm{aux}} = v_{\theta}\!\left(\mathbf{A}_t^{0}, \mathbf{z}_t\right)$ without the prior.
This pass yields the action expert's final-layer hidden states:
{
\setlength{\abovedisplayskip}{3pt}
\setlength{\belowdisplayskip}{3pt}
\setlength{\abovedisplayshortskip}{0pt}
\setlength{\belowdisplayshortskip}{0pt}
\begin{align}
    \mathbf{F}_t
    =
    \left(
        \mathbf{f}_t,\ldots,\mathbf{f}_{t+H-1}
    \right)
    \in \mathbb{R}^{H\times d_a},
    \label{eq:aux_denoising}
\end{align}
}%
where $d_a$ denotes the hidden dimension of the action expert,
and $\mathbf{f}_{t+j}$ is the final-layer hidden state corresponding to action $\mathbf{a}_{t+j}$ in the chunk.
This pass yields the action expert's coarse estimate of the upcoming trajectory given the observation,
aligned with each action position in the chunk.
We also train this pass with the flow-matching objective, so that it remains a valid denoising step with informative hidden states.
The auxiliary pass is used only to obtain $\mathbf{F}_t$ and its velocity for $\mathcal{L}_{\mathrm{aux}}$.

\vspace{-0.2cm}
\paragraph{Transition-timing Prediction Head.}
The prediction head estimates the transition-timing prior $\hat{\mathbf{p}}_t$
from the final-layer action-token features $\mathbf{F}_t$ and the cached
final-layer VLM features $\hat{\mathbf{z}}_t$ in $\mathbf{z}_t$.
It applies cross-attention between these features, followed by self-attention
over the action tokens, a linear projection
$\mathbf{w}_p\in\mathbb{R}^{d_z\times 1}$, and a sigmoid function $\sigma(\cdot)$:
{
\setlength{\abovedisplayskip}{3pt}
\setlength{\belowdisplayskip}{3pt}
\setlength{\abovedisplayshortskip}{0pt}
\setlength{\belowdisplayshortskip}{0pt}
\begin{equation}
\begin{aligned}
    \hat{\mathbf{p}}_t
    &= \sigma\!\left(\widetilde{\mathbf{F}}_t \mathbf{w}_p\right)
    \in [0,1]^{H}, \quad \widetilde{\mathbf{F}}_t = \bar{\mathbf{F}}_t + \operatorname{SelfAttn}(\bar{\mathbf{F}}_t) 
    \quad \text{where,~} \\
    \bar{\mathbf{F}}_t
    &= \hat{\mathbf{F}}_t + \operatorname{CrossAttn}\!\left(\hat{\mathbf{F}}_t,\, \hat{\mathbf{z}}_t,\, \hat{\mathbf{z}}_t\right)
    \in \mathbb{R}^{H\times d_z},
    \quad \hat{\mathbf{F}}_t = \mathbf{F}_t \mathbf{W}_f \in \mathbb{R}^{H\times d_z}.
\end{aligned}
\label{eq:transition_prior}
\end{equation}
}%
Here, 
$\mathbf{W}_f \in \mathbb{R}^{d_a \times d_z}$ is the projection matrix, and $d_z$ is the dimension of $\hat{\mathbf{z}}_t$.
The resulting transition-timing prior $\hat{\mathbf{p}}_t = (\hat{p}_{t}, \ldots, \hat{p}_{t+H-1}) \in [0,1]^{H}$ assigns a score to each action in the chunk, indicating how close it is to a subskill transition.
This prior then conditions action generation (Sec.~\ref{sec:cond}).

\vspace{-0.2cm}
\paragraph{Transition-timing Supervision.}
We derive the transition-timing target from demonstration actions as change points of PELT-based
unsupervised subskill discovery~\citep{cotpc,pelt} (Appendix~\ref{app:change_point_detection}).
Since a transition spans several steps and the detected points may not exactly match the true transition steps,
we convert each transition point into a soft target $\mathbf{p}_t$ that peaks at the point
and decays with distance (Appendix~\ref{app:soft_label}; Fig.~\ref{fig:prior_example}).
The head is then optimized with a binary cross-entropy loss
between the predicted prior $\hat{\mathbf{p}}_t$ and a target $\mathbf{p}_t=(p_{t},\ldots,p_{t+H-1})\in[0,1]^{H}$:
{
\setlength{\abovedisplayskip}{3pt}
\setlength{\belowdisplayskip}{3pt}
\setlength{\abovedisplayshortskip}{0pt}
\setlength{\belowdisplayshortskip}{0pt}
\begin{align}
    \mathcal{L}_{\mathrm{timing}}
    =
        \frac{1}{|\mathcal{T}|}
    \sum_{t\in\mathcal{T}}
    \operatorname{BCE}\!\left(\hat{\mathbf{p}}_t,\; \mathbf{p}_t\right),
    \label{eq:timing_loss}
\end{align}
}%
where $\mathcal{T}$ denotes the set of chunk start (\ie, policy call) steps across the demonstration trajectories.
Since the head reads the action-expert features from the auxiliary pass,
$\mathcal{L}_{\mathrm{timing}}$ is also back-propagated into the action expert,
encouraging its representations to capture upcoming transitions.

\vspace{-0.3cm}
\subsection{Transition-conditioned Action Generation}
\vspace{-0.2cm}
\label{sec:cond}
We now describe how RACE conditions the action expert on the transition-timing prior $\hat{\mathbf{p}}_t$ 
during the full denoising process in Eq.~(\ref{eq:race_denoising}) of Sec.~\ref{sec:overview}.
At each denoising step, the prior is injected through the action expert's adaptive RMSNorm layers~\citep{pi05}, 
which follow the adaptive normalization of DiT~\citep{dit},
as described below.

\input{tab/vlabench_comparison.tex}

\vspace{-0.25cm}
\paragraph{Transition-conditioned Modulation.}
To inject the timing prior $\hat{\mathbf{p}}_t\in[0,1]^H$ into the action expert,
we first construct token-wise transition features $\mathbf{U}_t^{k}\in \mathbb{R}^{H \times d_a}$ as:
{
\setlength{\abovedisplayskip}{3pt}
\setlength{\belowdisplayskip}{3pt}
\setlength{\abovedisplayshortskip}{0pt}
\setlength{\belowdisplayshortskip}{0pt}
\begin{align}
    \mathbf{U}_t^{k} = \alpha^{k}\, \hat{\mathbf{p}}_t \otimes \mathbf{e}_{\mathrm{trans}}
    \in \mathbb{R}^{H \times d_a},
    \label{eq:cue}
\end{align}
}%
where $\mathbf{e}_{\mathrm{trans}}\in\mathbb{R}^{d_a}$ is a randomly initialized learnable transition embedding,
$\otimes$ denotes the outer product,
and $\alpha^{k}$ is the learnable denoising-step gate from Eq.~(\ref{eq:race_denoising}),
which controls the strength of transition-prior conditioning at each step $k$.
We then map these features to token-wise offsets $[\Delta\boldsymbol{\Gamma}_\ell,\Delta\mathbf{B}_\ell,\Delta\mathbf{G}_\ell]$
of the scale, shift, and gate parameters of adaptive RMSNorm layer $\ell$,
and add them to the original modulation parameters, $\boldsymbol{\gamma}_\ell,\boldsymbol{\beta}_\ell,\mathbf{g}_\ell\in\mathbb{R}^{d_a}$, of the pretrained action expert:
{
\setlength{\abovedisplayskip}{3pt}
\setlength{\belowdisplayskip}{3pt}
\setlength{\abovedisplayshortskip}{0pt}
\setlength{\belowdisplayshortskip}{0pt}
\begin{equation}
\begin{aligned}
\left[\Delta\boldsymbol{\Gamma}_\ell, \Delta\mathbf{B}_\ell, \Delta\mathbf{G}_\ell\right]
&= \mathbf{U}_t^{k}\mathbf{W}_\ell \;\in\; \mathbb{R}^{H \times 3d_a}, \\
\left[\boldsymbol{\Gamma}_\ell, \mathbf{B}_\ell, \mathbf{G}_\ell\right]
&= \mathbf{1}_H\left[\boldsymbol{\gamma}_\ell^{\top}, \boldsymbol{\beta}_\ell^{\top}, \mathbf{g}_\ell^{\top}\right]
 + \left[\Delta\boldsymbol{\Gamma}_\ell, \Delta\mathbf{B}_\ell, \Delta\mathbf{G}_\ell\right],
\end{aligned}
\label{eq:token_modulation}
\end{equation}
}%
where $\mathbf{1}_H\in\mathbb{R}^{H\times1}$ is the all-ones vector that copies the original parameters to every action token,
and $\mathbf{W}_\ell\in\mathbb{R}^{d_a\times 3d_a}$ projects the transition features into token-wise offsets.
We initialize $\mathbf{W}_\ell$ to zero, so that training starts from the original modulation,
and $\mathbf{W}_\ell$ has no bias, so that a zero prior leaves the action expert unmodified.
These token-wise parameters $[\boldsymbol{\Gamma}_\ell,\mathbf{B}_\ell,\mathbf{G}_\ell]$ then modulate 
both the attention and feed-forward modules at each adaptive RMSNorm layer $\ell$ of the action expert:
{
\setlength{\abovedisplayskip}{3pt}
\setlength{\belowdisplayskip}{3pt}
\setlength{\abovedisplayshortskip}{0pt}
\setlength{\belowdisplayshortskip}{0pt}
\begin{align}
        \mathbf{X}_{\ell+1}
    = \mathbf{X}_\ell + \mathbf{G}_\ell \odot \mathrm{Module}_\ell(\tilde{\mathbf{X}}_\ell), \quad
    \tilde{\mathbf{X}}_\ell
    = (1+\boldsymbol{\Gamma}_\ell) \odot \mathrm{RMSNorm}(\mathbf{X}_\ell) + \mathbf{B}_\ell,
    \label{eq:adarms}
\end{align}
}%
where $\mathrm{Module}_\ell$ denotes either an attention or feed-forward module at layer $\ell$,
$\mathbf{X}_\ell \in \mathbb{R}^{H \times d_a}$ denotes the hidden states of the $H$ action tokens at layer $\ell$,
and $\odot$ denotes the element-wise product.
This conditioning mechanism allows the action expert to adapt its denoising behavior based on the predicted transition-timing prior $\hat{\mathbf{p}}_t$.

\vspace{-0.25cm}
\paragraph{Training Objective.}
Our framework is trained with the standard flow-matching objective~\citep{lipman2023flow} for both the full and auxiliary denoising passes,
together with the transition-timing prediction loss in Eq.~(\ref{eq:timing_loss}):
$\mathcal{L} = \mathcal{L}_{\mathrm{full}} + \lambda_{\mathrm{aux}}\mathcal{L}_{\mathrm{aux}} + \lambda_{\mathrm{timing}}\mathcal{L}_{\mathrm{timing}}$.
Here, both losses regress the velocity field toward the direction from the noise sample to the ground-truth chunk:
$\mathcal{L}_{\mathrm{full}}$ at a sampled flow time,
and $\mathcal{L}_{\mathrm{aux}}$ at the initial noise without modulation, \ie, on the auxiliary pass.
The coefficients $\lambda_{\mathrm{aux}}$ and $\lambda_{\mathrm{timing}}$ balance the two additional terms.
During training, the action expert is conditioned on the jittered target $\tilde{\mathbf{p}}_t$ rather than the prediction $\hat{\mathbf{p}}_t$
(\ie, teacher forcing and jittering; Appendix~\ref{app:effect_teacher_forcing_jittering}),
while at inference it is conditioned on the predicted $\hat{\mathbf{p}}_t$, as in Eq.~(\ref{eq:race_denoising}).

%% file: tab/vlabench_comparison.tex
\begin{table}[t]
\small
\centering
\caption{Comparison on VLABench. We report the success rate (SR) and the progress score (PS), the fraction of completed sub-tasks.
Spd is the per-episode inference speedup relative to the $\pi_{0.5}$ baseline (Appendix~\ref{app:speedup_measure}).
$H_{\text{exec}}$ is the number of actions executed per policy call, and
$\dagger$ denotes a model fine-tuned to execute that many actions.
$\ddagger$ indicates the average execution length, since adaptive chunking methods truncate each chunk adaptively.
Results for a wider range of chunk lengths are in Appendix~\ref{app:scaling_longer_chunks}.
}
\vspace{-0.15cm}
\label{tab:vlabench}
\setlength{\tabcolsep}{3pt}
\resizebox{\textwidth}{!}{%
\begin{tabular}{l|cc|cc|cc|cc|cc|cc|cc}
\hline
\multirow{2}{*}{Method} & \multirow{2}{*}{$H_{\text{exec}}$} & \multirow{2}{*}{Spd$\uparrow$}
& \multicolumn{2}{c|}{In-dist.} & \multicolumn{2}{c|}{Category}
& \multicolumn{2}{c|}{Common.} & \multicolumn{2}{c|}{Instruct.}
& \multicolumn{2}{c|}{Texture} & \multicolumn{2}{c}{Avg.} \\
& & & SR & PS & SR & PS & SR & PS & SR & PS & SR & PS & SR & PS \\
\hline
$\pi_{0.5}$ {\scriptsize Baseline}
    & 5 & $1.00\times$ & 40.4 & 56.7 & 21.4 & 35.6 & 17.0 & 33.7 & 18.0 & 35.9 & 26.0 & 42.1 & 24.6 & 40.8 \\ \hdashline[0.2pt/1pt]
RACE (ours)    & 5 & $0.96\times$ & 51.6 & 68.2 & 25.0 & 38.1 & 26.0 & 40.0 & 20.8 & 37.6 & 29.0 & 45.3 & 30.5 & 45.8 \\
\hline
\multicolumn{15}{c}{\textit{Action-Chunk Extension}} \\
\hdashline[0.2pt/1pt]
\multirow{3}{*}{$\pi_{0.5}$ {\scriptsize Fine-tuning}}
    & 10 & $2.01\times$ & 53.0 & 67.8 & 30.2 & 41.8 & 24.0 & 40.5 & 22.0 & 40.4 & 31.8 & 49.0 & 32.2 & 47.9 \\
    & 15 & $2.83\times$ & 51.4 & 65.6 & 26.6 & 39.6 & 23.4 & 40.3 & 19.0 & 37.1 & 29.4 & 46.3 & 30.0 & 45.8 \\
    & 20 & $3.82\times$ & 48.8 & 65.3 & 22.8 & 37.2 & 20.2 & 38.2 & 20.8 & 39.3 & 27.8 & 43.9 & 28.1 & 44.8 \\
\hdashline[0.2pt/1pt]
\multirow{3}{*}{RACE (ours)}
    & 10 & $1.92\times$ & 57.6 & 70.9 & 30.4 & 41.8 & 23.2 & 39.0 & 25.0 & 42.0 & 36.0 & 52.8 & 34.4 & 49.3 \\
    & 15 & $2.71\times$ & 54.2 & 67.1 & 28.8 & 39.4 & 25.6 & 41.4 & 25.2 & 42.6 & 33.2 & 50.0 & 33.4 & 48.1 \\
    & 20 & $3.68\times$ & 54.0 & 68.2 & 27.2 & 38.8 & 28.2 & 41.4 & 21.0 & 39.4 & 34.8 & 51.0 & 33.0 & 47.8 \\
\hline
\multicolumn{15}{c}{\textit{Recent State-of-the-Art VLAs}} \\
\hdashline[0.2pt/1pt]
\multirow{2}{*}{ACoT-VLA~\citep{acotvla}}
    & 5  & 0.62$\times$ & 46.4 & 68.0 & 22.2 & 40.0 & 24.4 & 42.0 & 20.4 & 39.2 & 34.4 & 55.2 & 29.6 & 48.9 \\
    & 20$^{\dagger}$  & 2.36$\times$ & 49.4 & 70.8 & 21.4 & 39.2 & 24.4 & 43.0 & 28.2 & 46.0 & 42.0 & 65.2 & 33.1 & 52.8 \\
    \hline
    \multicolumn{15}{c}{\textit{Adaptive Chunking Methods}} \\ \hdashline[0.2pt/1pt]
    AutoHorizon~\citep{autohorizon} & 9.45$^{\ddagger}$ & 1.89$\times$ & 43.8 & 59.5 & 23.8 & 37.8 & 16.4 & 33.3 & 18.2 & 36.3 & 24.4 & 40.6 & 25.3 & 41.5 \\
PACE~\citep{pace} & 8.58$^{\ddagger}$ & $1.69\times$ & 41.2 & 58.4 & 22.8 & 36.4 & 20.0 & 36.8 & 16.0 & 35.6 & 26.8 & 44.4 & 25.4 & 42.3 \\
\hline
    \hline
\multicolumn{15}{c}{\textit{Other Efficient VLA Methods}} \\ \hdashline[0.2pt/1pt]
Latent Bridge~\citep{latentbridge} & 5 & $1.63\times$ & 39.2 & 56.8 & 23.2 & 37.0 & 19.4 & 36.0 & 14.8 & 33.2 & 28.0 & 44.8 & 24.9 & 41.6 \\
GridS~\citep{grids} & 5 & $1.24\times$ & 46.2 & 63.4 & 24.0 & 38.4 & 19.8 & 38.4 & 13.6 & 32.0 & 38.4 & 55.2 & 28.4 & 45.5 \\
\hline
\end{tabular}
}
\vspace{-0.2cm}
\end{table}

%% file: sec/05_experiment.tex
\vspace{-0.3cm}
\section{Experiments}
\vspace{-0.3cm}

\subsection{Experimental Setup}
\label{sec:setup}
\vspace{-0.2cm}

\paragraph{Dataset and Metric.}
For simulation experiments, we evaluate our framework on three manipulation benchmarks: 
VLABench~\citep{vlabench}, RoboCasa-H50~\citep{robocasa}, and LIBERO~\citep{libero},
with the LIBERO results reported in Appendix~\ref{app:comparison_on_libero}.
We follow the standard evaluation protocols of each benchmark, with the dataset and protocol details provided in Appendix~\ref{app:dataset}.
Throughout, $H_{\text{exec}}$ denotes the number of actions executed per policy call.
For each $H_{\text{exec}}$, we post-train RACE on top of the $\pi_{0.5}$ fine-tuned for that benchmark,
together with a fine-tuned $\pi_{0.5}$ variant trained with the same recipe;
both predict and execute a full chunk of $H_{\text{exec}}$ actions.
The original $\pi_{0.5}$ instead predicts $10$ actions and executes only the first $5$,
following its official protocol.
As evaluation metrics, we measure the task success rate and the speedup relative to the baseline $\pi_{0.5}$,
measured as the reduction in cumulative policy-inference time per episode, excluding simulator execution time
(Appendix~\ref{app:speedup_measure}).
The real-world setup and metrics are in Appendix~\ref{app:real-world}.

\vspace{-0.2cm}
\paragraph{Implementation Details.}
We build our framework on the official $\pi_{0.5}$~\citep{pi05} checkpoints fine-tuned for each benchmark,
with $K=10$ denoising steps.
For the transition-timing prediction (Sec.~\ref{sec:head}),
the cached VLM features and the action-expert hidden states have dimensions
$d_{z}=256$ and $d_{a}=1024$.
For the transition-conditioned generation (Sec.~\ref{sec:cond}),
we initialize every gate $\alpha^{k}$ to $0.5$
and optimize it during training.
In the training objective, we set $\lambda_{\mathrm{aux}}=0.1$ and $\lambda_{\mathrm{timing}}=0.05$.
We train for 40K steps on VLABench and 20K steps on the other benchmarks
using AdamW with a learning rate that decays from $5\times10^{-5}$ to $5\times10^{-6}$ following a cosine schedule,
and a batch size of 64 on four NVIDIA RTX A6000 GPUs.

\vspace{-0.3cm}
\subsection{Main Results}
\vspace{-0.2cm}

\input{tab/robocasa_h50.tex}

\paragraph{Comparison on VLABench.}
Tab.~\ref{tab:vlabench} reports the success rate (SR), the progress score (PS), \ie, the fraction of completed sub-tasks,
and the per-episode inference speedup across the five VLABench tracks.
At every $H_{\text{exec}}\in\{10,15,20\}$, RACE outperforms the fine-tuned $\pi_{0.5}$ executing the same number of actions,
and at $H_{\text{exec}}=20$, it still outperforms the $\pi_{0.5}$ baseline while requiring far less inference time per episode.
At $H_{\text{exec}}=5$, RACE also achieves a higher success rate than the official $\pi_{0.5}$ checkpoint.
ACoT-VLA~\citep{acotvla}, a recent state-of-the-art VLA built on $\pi_{0.5}$,
reaches a success rate comparable to RACE at $H_{\text{exec}}=20$ with a higher progress score but a lower speedup,
as it runs an additional multi-step denoising process whereas RACE adds a single auxiliary pass
(Appendix~\ref{app:latency_overhead} and~\ref{app:training_efficiency}).
With $H_{\text{exec}}\geq10$, RACE also outperforms adaptive chunking methods~\citep{autohorizon,pace}
and efficient VLA methods~\citep{latentbridge,grids} in both success rate and speedup.
Results for longer chunks are in Appendix~\ref{app:scaling_longer_chunks}.

\input{tab/ablation.tex}
\input{tab/design_choices.tex}

\begin{figure}[t]
    \centering
    \begin{subfigure}[t]{0.24\linewidth}
        \centering
        \includegraphics[width=\linewidth]{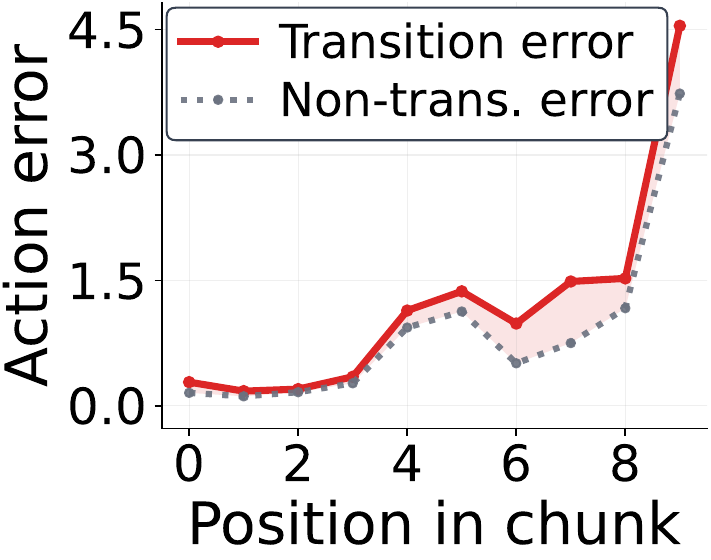}
        \vspace{-0.5cm}
        \caption{$H_{\text{exec}}=10$; $\pi_{0.5}$}
        \label{fig:drift_error_h10_ft}
    \end{subfigure}\hfill
    \begin{subfigure}[t]{0.24\linewidth}
        \centering
        \includegraphics[width=\linewidth]{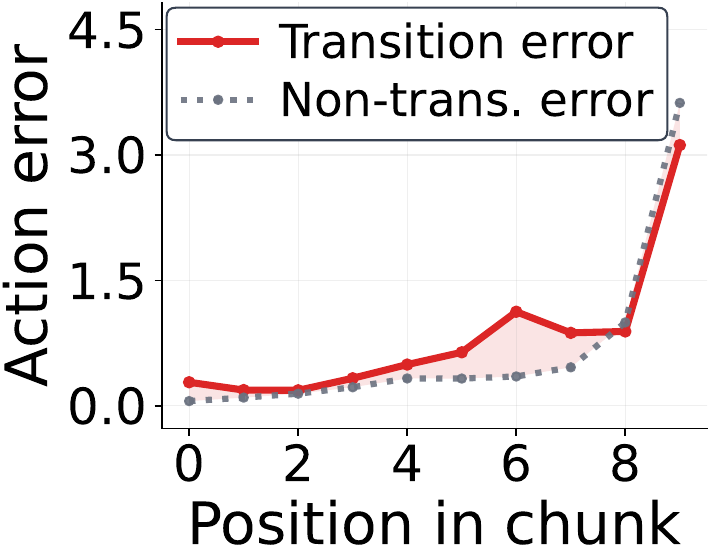}
        \vspace{-0.5cm}
        \caption{$H_{\text{exec}}=10$; ours}
        \label{fig:drift_error_h10_ours}
    \end{subfigure}\hfill
    \begin{subfigure}[t]{0.24\linewidth}
        \centering
        \includegraphics[width=\linewidth]{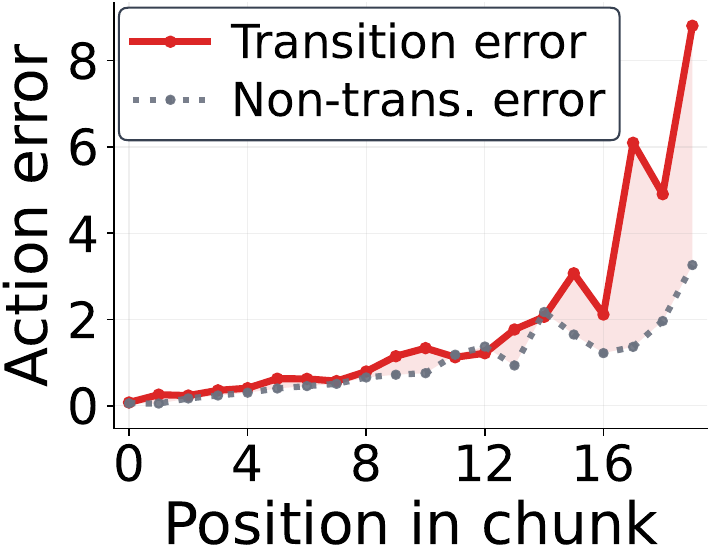}
        \vspace{-0.5cm}
        \caption{$H_{\text{exec}}=20$; $\pi_{0.5}$}
        \label{fig:drift_error_h20_ft}
    \end{subfigure}\hfill
    \begin{subfigure}[t]{0.24\linewidth}
        \centering
        \includegraphics[width=\linewidth]{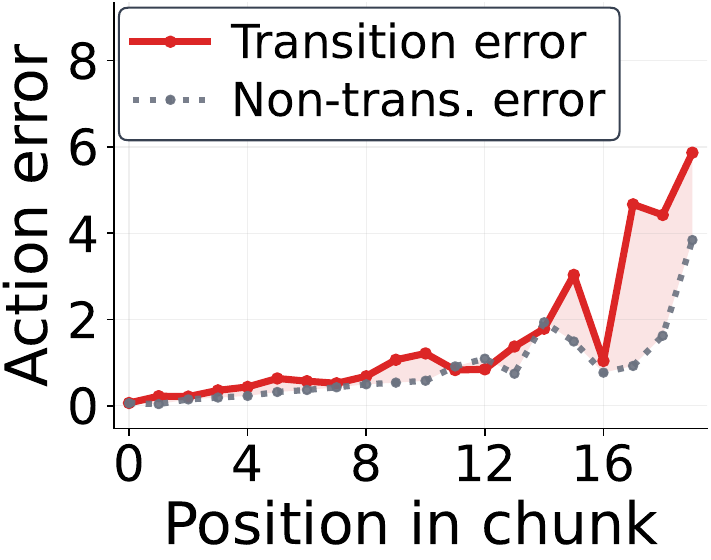}
                \vspace{-0.5cm}
        \caption{$H_{\text{exec}}=20$; ours}
        \label{fig:drift_error_h20_ours}
    \end{subfigure}
            \vspace{-0.2cm}
    \caption{
        Errors at transition and non-transition actions with $H_{\text{exec}}=10$ and $20$ on VLABench, 
    where $\pi_{0.5}$ is fine-tuned to execute $H_{\text{exec}}$ actions per policy call.
    At each position within a chunk, we compare the error of actions at transition points (transition error)
with that of the remaining actions (non-transition error), which we use as a reference for the error growth with prediction horizon.
    }
            \vspace{0.2cm}
    \label{fig:drift_error}
    \begin{subfigure}[t]{0.32\textwidth}
        \centering
        \includegraphics[scale=0.46]{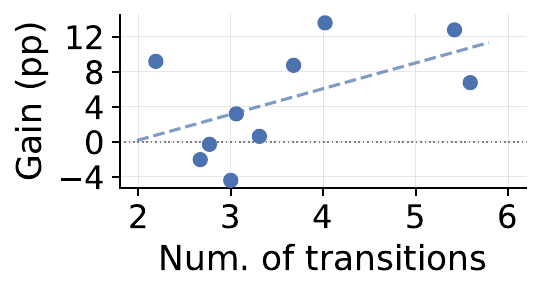}
        \vspace{-0.2cm}
        \caption{Gain on \# of transitions}
        \label{fig:gain_vs_transitions}
    \end{subfigure}
    \hfill
    \begin{subfigure}[t]{0.32\textwidth}
        \centering
        \includegraphics[scale=0.46]{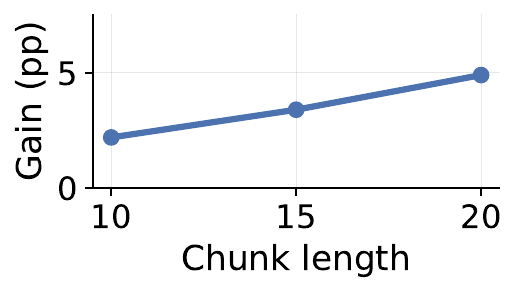}
        \vspace{-0.2cm}
        \caption{Gain on chunk lengths}
        \label{fig:gain_vs_chunk}
    \end{subfigure}
    \hfill
    \begin{subfigure}[t]{0.32\textwidth}
        \centering
        \includegraphics[scale=0.46]{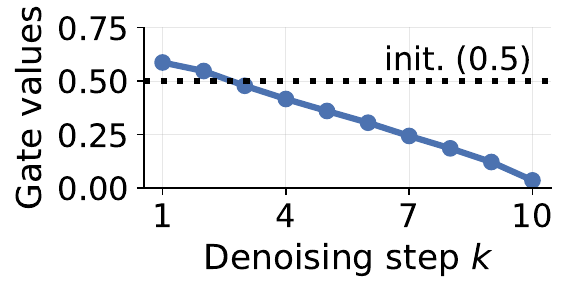}
        \vspace{-0.64cm}
        \caption{Learned gate values $\alpha^{k}$}
        \label{fig:learned_gates}
    \end{subfigure}
    \vspace{-0.2cm}
    \caption{Effect of the transition-timing prior on VLABench.
    (a) gain of RACE over the fine-tuned $\pi_{0.5}$ as a function of the number of transitions in each task on VLABench,
    (b) the same gain of RACE at each chunk length,
    and (c) learned gate values $\alpha^{k}$ in Eq.~(\ref{eq:race_denoising}) across denoising steps.
    }
    \vspace{-0.5cm}
    \label{fig:prior_analysis_figs}
\end{figure}

\vspace{-0.25cm}
\paragraph{Comparison on RoboCasa-H50.}
Tab.~\ref{tab:robocasa_h50} reports success rates and per-episode inference speedup on the eight RoboCasa-H50 categories,
with the average weighted by the number of tasks per category.
RACE outperforms the fine-tuned $\pi_{0.5}$ at every $H_{\text{exec}}$,
and at $H_{\text{exec}}=20$, it still outperforms the $\pi_{0.5}$ baseline while requiring far less inference time per episode.
At $H_{\text{exec}}=5$, RACE also achieves a higher success rate than the official $\pi_{0.5}$ checkpoint.
The recent state-of-the-art VLA, ACoT-VLA~\citep{acotvla}, fine-tuned to execute $H_{\text{exec}}=20$ actions falls below RACE at the same length, with a lower speedup.
With $H_{\text{exec}}\geq10$, RACE also outperforms efficient VLA methods~\citep{latentbridge,grids} in both success rate and speedup,
and adaptive chunking methods~\citep{autohorizon,pace} at $H_{\text{exec}}\in\{10,15\}$,
while remaining comparable at $H_{\text{exec}}=20$ with about twice their speedup.

\paragraph{Comparison on LIBERO.}
On LIBERO, RACE outperforms the fine-tuned $\pi_{0.5}$ at every $H_{\text{exec}}$,
and PolicyTrim~\citep{policytrim} in success rate with $H_{\text{exec}}\in\{10,15\}$ (Appendix~\ref{app:comparison_on_libero}).

\vspace{-0.2cm}
\subsection{Ablation Study}
\vspace{-0.2cm}

We conduct ablation studies with an execution chunk length of $H_{\text{exec}}=20$ on VLABench. 

\vspace{-0.25cm}
\paragraph{Effect of Each Component and Training Objective.}
In Tab.~\ref{tab:ablation}, we add each component and training objective one at a time, 
starting from the fine-tuned $\pi_{0.5}$ with only $\mathcal{L}_{\text{full}}$.
Adding $\mathcal{L}_{\mathrm{aux}}$ alone, without the prediction head, yields only a small improvement,
so the extra denoising objective contributes little by itself.
Adding the prediction head with $\mathcal{L}_{\mathrm{timing}}$ raises the success rate,
even though the prediction is not injected into generation.
Since the head reads the action-expert features from the auxiliary pass,
$\mathcal{L}_{\mathrm{timing}}$ back-propagates into the action expert and makes its features transition-aware.
Conditioning on the predicted prior improves it further under the same training objectives,
showing that timing supervision and timing conditioning provide complementary gains.
Finally, $\mathcal{L}_{\mathrm{aux}}$ keeps the auxiliary pass a valid denoiser,
which may provide better features for the head.

\vspace{-0.25cm}
\paragraph{Ablation within Transition-timing Prediction Head.}
In Tab.~\ref{tab:design_choices},
we conduct the ablation study on the transition labels and input information sources 
for the transition-timing prediction head (Sec.~\ref{sec:head}).
\textbf{(a) Transition labels} (Tab.~\ref{tab:design_choices_labels}).
Replacing the PELT labels with the random transitions shrinks the learned modulation
offsets by more than an order of magnitude;
the model learns to ignore the uninformative prior and falls back to na\"ive fine-tuning.
Speed-minimum labels~\citep{pace} also underperform PELT.
Appendix~\ref{app:validity_transition_labels} compares these label sets against the transitions of the demonstration generator,
and less accurate labels indeed yield smaller gains.
Further results with other PELT settings are reported in Appendix~\ref{app:sensitivity_pelt}.
\textbf{(b) Information sources} (Tab.~\ref{tab:design_choices_information}).
Using the VLM features alone performs worst:
with the head attached to the VLM, $\mathcal{L}_{\mathrm{timing}}$ never reaches the action expert,
so the gain from reshaping its features (Tab.~\ref{tab:ablation}) is lost.
The action-expert features alone come second, and combining both performs best
(further ablations are in Appendix~\ref{appendix:additional_ablation}).

\vspace{-0.25cm}
\subsection{Understanding the Effectiveness of RACE}
\label{subsec:understanding}

\vspace{-0.15cm}
\paragraph{Where Errors Arise in Longer Chunks.}
To examine where errors arise as chunks grow longer,
we compare actions predicted from demonstration states with the demonstrated actions
on VLABench at $H_{\text{exec}}\in\{10,20\}$
(Fig.~\ref{fig:drift_error}; Appendix~\ref{app:action_error_measure}).
We compare errors at transitions with those of non-transition actions at the same chunk position.
Both errors grow toward the end of the chunk, as expected from open-loop prediction farther ahead.
However, errors at transitions remain generally higher, and this gap widens at $H_{\text{exec}}=20$.
This indicates that transitions pose additional difficulty beyond the growth with prediction horizon.
RACE reduces errors at transitions over the fine-tuned $\pi_{0.5}$, particularly near the end of the chunk,
while also lowering errors at non-transition actions.
For the gripper, whose binary command mainly reflects when it switches,
RACE switches within one step of the correct step more often and misses fewer switches than the fine-tuned $\pi_{0.5}$,
with a similar false-switch rate
(Tab.~\ref{tab:gripper_timing} of Appendix~\ref{app:action_errors_dimension}, with transition errors by action component in Tab.~\ref{tab:transition_error_by_component}).

\vspace{-0.22cm}
\paragraph{Gains on Transition-rich Tasks.}
Fig.~\ref{fig:gain_vs_transitions} illustrates the gain of RACE over the fine-tuned $\pi_{0.5}$ at $H_{\text{exec}}=20$ for ten VLABench tasks.
For each task, gains are averaged over five tracks and plotted against the average number of transitions per training demonstration.
Although gains vary across tasks, RACE tends to achieve larger improvements
on tasks with more subskill transitions.
This trend is consistent with our motivation to improve action generation
around subskill transitions.
The detailed statistics of each task are provided in Appendix~\ref{app:statistics_transitions}.

\vspace{-0.22cm}
\paragraph{Gains on Longer Chunks.}
Fig.~\ref{fig:gain_vs_chunk} shows the gain of RACE over the fine-tuned $\pi_{0.5}$ across chunk lengths on VLABench.
The gain is positive at every $H_{\text{exec}}$ and grows as the chunk becomes longer.
Since a longer chunk is more likely to span one or more subskill transitions,
this pattern indicates that transition timing is an obstacle to chunk extension
and that RACE relieves it.

\begin{wrapfigure}{r}{0.55\linewidth}
    \vspace{-9pt}
    \centering
    \begin{subfigure}[t]{0.455\linewidth}
        \centering
        \includegraphics[width=\linewidth]{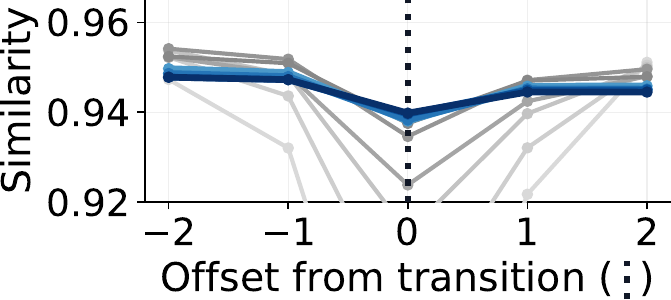}
        \vspace{-0.5cm}
        \caption{$\pi_{0.5}$}
        \label{fig:emergence_a}
    \end{subfigure}\hfill
    \begin{subfigure}[t]{0.525\linewidth}
        \centering
        \includegraphics[width=\linewidth]{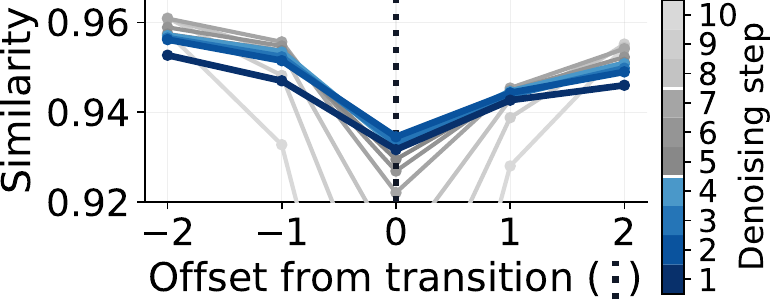}
        \vspace{-0.5cm}
        \caption{RACE (ours)}
        \label{fig:emergence_b}
    \end{subfigure}
    \vspace{-0.3cm}
    \caption{Adjacent action-token similarity.}
    \label{fig:emergence}
    \vspace{-10pt}
\end{wrapfigure}

\vspace{-0.22cm}
\paragraph{Learned Gates across Denoising Steps.}
In Fig.~\ref{fig:learned_gates}, we report the learned per-step gates $\alpha^{k}$ of Eq.~(\ref{eq:race_denoising}),
which control how strongly the prior modulates each denoising step.
Initialized at the same value, the gates decrease monotonically across denoising steps after training,
so the prior modulates the early steps most strongly.
This pattern is in line with our analysis in Fig.~\ref{fig:emergence}.

\vspace{-0.22cm}
\paragraph{Transition Boundaries across Denoising Steps.}
To examine how RACE changes the action expert's representations,
we measure the cosine similarity between the hidden states of adjacent action tokens, aligned at the transition point,
at each denoising step in Fig.~\ref{fig:emergence}.
A dip at the transition indicates that the action expert separates the two neighboring behaviors.
In $\pi_{0.5}$, the dip is weak at the early steps and becomes clearer only later.
With RACE, the dip is more pronounced from the first step,
consistent with the timing supervision and the learned gates (Fig.~\ref{fig:learned_gates}), which peak early.

\vspace{-0.22cm}
\begin{wraptable}{r}{0.48\linewidth}
\vspace{-9pt}
\small
\centering
\caption{Replacement and shift of timing prior.}
\vspace{-6.8pt}
\label{tab:inference_prior_perturbing_replacement_wrap}
\setlength{\tabcolsep}{2.7pt}
\scalebox{0.99}{
\begin{tabular}{l|ccc|cc}
\hline
& \multicolumn{3}{c|}{Timing prior at inference} & \multicolumn{2}{c}{Prior shift} \\
\cline{2-6}
& Random & Zero & Predicted (ours) & $-1$ & $+1$ \\
\hline
Avg. & 23.5 & 31.8 & {33.0} & 30.2 & 30.7 \\
\hline
\end{tabular}
}
\vspace{-8pt}
\end{wraptable}
\paragraph{Perturbing Transition-timing at Inference.}
To examine how much the predicted timing contributes at inference,
we keep the trained model fixed and replace the prior with zeros or random peaks, or shift it by one step
(Tab.~\ref{tab:inference_prior_perturbing_replacement_wrap}; VLABench, $H_{\text{exec}}=20$).
A zero prior removes the modulation and lowers the success rate,
showing that conditioning contributes at inference.
Much of the gain over the fine-tuned $\pi_{0.5}$ remains,
consistent with the transition-aware features learned through timing supervision (Tab.~\ref{tab:ablation}).
A random prior lowers it further.
Shifting the predicted prior by a single step also lowers the success rate below the zero-prior setting
and moves the generated gripper switches in the same direction (Appendices~\ref{app:sensitivity_to_shift} and~\ref{app:prior_generated_timing}),
showing that the model relies on where the prior places transitions, not merely on its presence.
Although the predicted prior does not capture every transition (Appendix~\ref{app:accuracy_transition_timing_prediction}),
it yields the highest success rate among these settings.

\vspace{-0.2cm}
\subsection{Real-World Deployment}
\label{subsec:real_world}

\vspace{-0.2cm}
\begin{wraptable}{r}{0.45\linewidth}
\vspace{-10pt}
\small
\centering
\caption{Results on a real robot.}
\vspace{-7pt}
\label{tab:realworld_pnp}
\setlength{\tabcolsep}{4pt}
\begin{tabular}{l|c|ccc}
\hline
Method & $H_{\text{exec}}$ & SR$\uparrow$ & Time$\downarrow$ & Idle$\downarrow$ \\
\hline
\multirow{2}{*}{$\pi_{0.5}$ {\scriptsize Fine-tuning}}
    & 5 & 36\% & 12.6\,s & 2.11\,s \\
    & 20 & 48\% & 11.6\,s & 0.40\,s \\
\hdashline[0.2pt/1pt]
\multirow{1}{*}{RACE (ours)}
    & 20 & 66\% & 11.8\,s & 0.41\,s \\
\hline
\end{tabular}
\vspace{-10pt}
\end{wraptable}
\paragraph{Results on a Real Robot.}
To verify that RACE reliably executes longer action chunks that reduce idle time on a real robot,
we fine-tune each policy on 30 demonstrations of a pick-and-place task and evaluate it over 50 trials
(details in Appendix~\ref{app:real-world}; demo videos on \href{https://github.com/Seonghoon-Yu/RACE-VLA}{GitHub}).
All methods use asynchronous execution with real-time chunking~\citep{rtc},
generating the next chunk while the current one is executed;
each policy thus predicts a chunk twice as long as it executes
(\eg, $H=40$ for $H_{\text{exec}}=20$).
We report the success rate, task-completion time, and idle time averaged over successful trials,
\ie, the total time the robot receives no new action command.
As shown in Tab.~\ref{tab:realworld_pnp}, a short chunk still leaves idle time even with asynchronous execution
(analyzed in Appendix~\ref{app:idle_time}).
Extending the chunk greatly reduces it,
since executing more actions per policy call leaves enough time to generate the next chunk;
the remaining idle time comes mostly from the first policy call at the start of each episode.
Task-completion time decreases only slightly,
since asynchronous execution already hides most of the latency
and the completion time is dominated by the robot motion itself;
the reduced idle time mainly removes pauses that interrupt smooth motion
and can shift the robot dynamics away from the demonstrations~\citep{rtc}.
At the same chunk length, RACE achieves a higher success rate than fine-tuning.
\leavevmode\par
\ifnum\value{WF@wrappedlines}>1\relax
    \vspace{\dimexpr\value{WF@wrappedlines}\baselineskip-\baselineskip\relax}
\fi
\WFclear

%% file: tab/robocasa_h50.tex
\begin{table}[t]
\small
\centering
\caption{Comparison on RoboCasa-H50. We report the success rate on eight task categories,
with the average weighted by the number of tasks per category.
Spd is the per-episode inference speedup relative to the $\pi_{0.5}$ baseline.
$\ddagger$ indicates the average execution length, since adaptive chunking methods truncate each chunk adaptively.
}
\vspace{-0.15cm}
\label{tab:robocasa_h50}
\setlength{\tabcolsep}{2.7pt}
\resizebox{0.95\textwidth}{!}{%
\begin{tabular}{l|cc|cccccccc|c}
\hline
Method & $H_{\text{exec}}$ & Spd$\uparrow$ & PnP & Doors & Drawer & Sink & Stove & Coffee & Micro. & Nav. & Avg. \\
\hline
$\pi_{0.5}$ {\scriptsize Baseline}
 & 5 & $1.00\times$ & 51.5 & 57.0 & 95.0 & 74.7 & 40.0 & 65.3 & 79.0 & 22.0 & 60.4 \\ \hdashline[0.2pt/1pt]
RACE (ours)  & 5 & $0.97\times$ & 57.2 & 56.0 & 93.0 & 77.3 & 42.0 & 68.0 & 80.0 & 40.0 & 63.5 \\
\hline
\multicolumn{12}{c}{\textit{Action-Chunk Extension}} \\
\hdashline[0.2pt/1pt]
\multirow{3}{*}{$\pi_{0.5}$ {\scriptsize Fine-tuning}}
    & 10 & 1.96$\times$ & 62.2 & 56.0 & 94.0 & 75.3 & 45.0 & 65.3 & 89.0 & 26.0 & 65.0 \\
    & 15 & 2.94$\times$ & 60.5 &	58.0 &	94.0 &	68.7 &	35.0 &	72.0 &	91.0 &	26.0 &	64.2 \\
    & 20 & 3.84$\times$ & 56.8 & 61.0 & 89.0 & 62.7 & 38.0 & 62.7 & 87.0 & 18.0 & 60.8 \\
\hdashline[0.2pt/1pt]
\multirow{3}{*}{RACE (ours)}
 & 10 & 1.91$\times$ & 65.0 & 61.0 & 95.0 & 73.3 & 42.0 & 70.0 & 84.0 & 34.0 & 66.8 \\
 & 15 & 2.91$\times$ & 61.2 & 65.5 & 95.0 & 79.3 & 44.0 & 69.3 & 92.0 & 24.0 & 67.3 \\
 & 20 & 3.77$\times$ & 62.2 & 63.5 & 93.0 & 68.0 & 34.0 & 64.7 & 84.0 & 28.0 & 64.0 \\
 \hline
\multicolumn{12}{c}{\textit{Recent State-of-the-Art VLAs}} \\
\hdashline[0.2pt/1pt]
\multirow{2}{*}{ACoT-VLA~\citep{acotvla}} & 5 & 0.70$\times$ & 58.2	&61.5	&89.0&	74.7&	40.0&	67.3&	92.0&	28.0&	64.3 \\
                         & 20 & 2.66$\times$ &55.0&	58.0&	79.0&	64.0&	23.0&	57.3&	75.0&	38.0&	57.1 \\
\hline
\multicolumn{12}{c}{\textit{Adaptive Chunking Methods}} \\ \hdashline[0.2pt/1pt]
AutoHorizon~\citep{autohorizon} & 9.41$^{\ddagger}$ & 1.87$\times$ & 58.5 & 58.0 & 90.0 & 74.7 & 34.0 & 64.7 & 84.0 & 42.0 & 63.0 \\
PACE~\citep{pace}               & 8.11$^{\ddagger}$ & 1.63$\times$ & 61.3 & 57.5 & 91.0 & 74.0 & 32.0 & 67.3 & 85.0 & 44.0 & 64.2 \\
\hline
    \hline
\multicolumn{12}{c}{\textit{Other Efficient VLA Methods}} \\ \hdashline[0.2pt/1pt]
Latent Bridge~\citep{latentbridge}       & 5 & 1.63$\times$  & 56.2 & 59.0 & 93.0 & 67.3 & 33.0 & 58.7 & 74.0 & 44.0 & 60.3 \\
GridS~\citep{grids}              & 5 & 1.25$\times$  & 46.5 & 50.5 & 90.0 & 71.3 & 26.0 & 57.3 & 70.0 & 46.0 & 55.1 \\
\hline
\end{tabular}
}
\vspace{-0.1cm}
\end{table}

%% file: tab/ablation.tex
\begin{table}[t]
\small
\centering
\caption{Ablation of each component and training objective.
The auxiliary one-step pass is always performed when the transition-timing prediction head is present, since the head reads its features;
$\mathcal{L}_{\mathrm{aux}}$ only controls whether this pass is additionally supervised as a denoiser.
}
\vspace{-0.2cm}
\label{tab:ablation}
\setlength{\tabcolsep}{6pt}
\scalebox{0.95}{
\begin{tabular}{cc|ccc|c}
\hline
\multirow{2}{*}{\makecell{Transition-timing\\Prediction Head (Sec.~\ref{sec:head})}}
& \multirow{2}{*}{\makecell{Transition-conditioned\\Action Generation (Sec.~\ref{sec:cond})}}
& \multicolumn{3}{c|}{Training Objectives}
& \multirow{2}{*}{Avg. SR} \\
\cline{3-5}
& & $\mathcal{L}_{\mathrm{full}}$
  & $\mathcal{L}_{\mathrm{aux}}$
  & $\mathcal{L}_{\mathrm{timing}}$ & \\
\hline
- & - & $\checkmark$ & - & - & 28.1 \\
  \hdashline[0.2pt/1pt]
  - & - & $\checkmark$ & $\checkmark$ & -  & 28.4 \\
  $\checkmark$ & - & $\checkmark$ & - & $\checkmark$  & 30.6 \\
$\checkmark$ & $\checkmark$ & $\checkmark$ & - & $\checkmark$ & 32.4 \\
\hdashline[0.2pt/1pt]
$\checkmark$ & $\checkmark$ & $\checkmark$ & $\checkmark$ & $\checkmark$ & 33.0 \\
\hline
\end{tabular}
}
\vspace{-0.5cm}
\end{table}

%% file: tab/design_choices.tex
\begin{table*}[t]
\small
\centering
\caption{Ablation within the transition-timing prediction head (Sec.~\ref{sec:head}). 
Analyses of transition-label validity and sensitivity to PELT settings are provided in
Appendices~\ref{app:validity_transition_labels} and~\ref{app:sensitivity_pelt}, respectively.
}
\vspace{-0.25cm}
\label{tab:design_choices}
\begin{subtable}[t]{0.665\textwidth}
\centering
\caption{Transition-label sources
}
\vspace{-0.1cm}
\label{tab:design_choices_labels}
\setlength{\tabcolsep}{6pt}
\fontsize{8}{9.5}\selectfont
\begin{tabular}{l|c||ccc}
\hline
\multirow{2}{*}{Label source} & \multirow{2}{*}{Avg.} & \multicolumn{3}{c}{Modulation offsets (Sec.~\ref{sec:cond})} \\
\cline{3-5}
& & ~$\|\Delta\boldsymbol{\Gamma}\|$~ & $\|\Delta\mathbf{B}\|$~ & $\|\Delta\mathbf{G}\|$~ \\
\hline
Random & 28.5 & 0.30 & 0.35 & 0.35 \\
Speed minima~\citep{pace} & 31.2 & 5.45 & 6.37 & 5.72 \\
\hdashline[0.2pt/1pt]
PELT (used) & 33.0 & 3.67 & 4.12 & 3.81 \\
\hline
\end{tabular}
\end{subtable}
\hfill
\begin{subtable}[t]{0.32\textwidth}
\centering
\caption{Information sources}
\vspace{-0.1cm}
\label{tab:design_choices_information}
\setlength{\tabcolsep}{7pt}
\fontsize{8}{9.5}\selectfont
\begin{tabular}{l|c}
\hline
\multirow{2}{*}{\makecell[l]{Input features for\\the head~(Sec.~\ref{sec:head})}} & \multirow{2}{*}{Avg.} \\
& \\
\hline
VLM features only & 31.6 \\
Action features only & 32.5 \\
\hdashline[0.2pt/1pt]
Both (ours) & 33.0 \\
\hline
\end{tabular}
\end{subtable}
\vspace{-0.1cm}
\end{table*}

%% file: sec/10_conclusion.tex
\vspace{-0.35cm}
\section{Conclusion}
\vspace{-0.35cm}

In this work, we study why longer action chunks become unreliable in VLAs,
and observe that action errors concentrate at transitions between manipulation subskills and grow with the chunk length.
Based on this, we propose RACE, which improves long-chunk execution
through transition-aware learning and transition-conditioned generation,
resulting in competitive success rates while executing $4\times$ longer action chunks.
On a real robot, RACE enables longer chunks that reduce idle time, with a higher success rate than fine-tuning at the same chunk length.

\vspace{-0.1cm}
\subsection*{AI use statement}
\vspace{-0.1cm}
In this work, we used generative AI tools to edit the paper for readability,
to draft and revise parts of the text, to search for related work,
to assist in implementing the method and experiments,
and to name the subskills in Fig.~\ref{fig:transition_examples} for illustration only,
which are not used for training or analysis.
All code was designed, reviewed, and tested by the authors,
and all reported results were verified by the authors.
We have not used generative AI tools for generating synthetic datasets, developing theoretical models,
formulating or proving mathematical claims, proposing or refining hypotheses,
designing the research methodology, translating content, cleaning datasets,
qualitative analysis, or interpreting results.
We have reviewed all AI-assisted content and take responsibility for the final content of this work,
including text, code, and claims produced with the aid of generative AI.

\vspace{-0.1cm}
\subsection*{Ethics statement}
\vspace{-0.1cm}
This work uses publicly available simulation benchmarks and demonstrations collected in our laboratory,
and does not involve human subjects or personal data.
Since longer action chunks reduce how often the robot reacts to new observations,
deploying such policies near people requires additional safeguards (Appendix~\ref{app:broader_impact}).

\vspace{-0.1cm}
\subsection*{Reproducibility statement}
\vspace{-0.1cm}
To support reproducibility, we will release the source code at 
\href{https://github.com/Seonghoon-Yu/RACE-VLA}{Github}.
It will include the training code for RACE and the fine-tuned $\pi_{0.5}$ baseline on VLABench,
the transition labels together with the script that computes them using PELT,
and instructions for running inference with a trained model.
The implementation details are described in Sec.~\ref{sec:setup} and the appendix,
including the transition discovery (Appendix~\ref{app:transition}),
the dataset and evaluation protocols (Appendix~\ref{app:dataset}),
and the pseudocode for training and inference (Appendix~\ref{app:pseudocode}).
We will also release the code for the other benchmarks,
together with the trained model weights.

%% file: sec/XX_appendix.tex
\clearpage

\appendix

\begin{center}
\toptitlebar
    {\Large \textbf{When to Switch: Reliable Action-Chunk Extension \\for Vision-Language-Action Models} \\ - \textit{Appendix} - \par}
\bottomtitlebar
\end{center}

\begin{center}
    \large{\textbf{Overview of Appendix}}
\end{center}
We provide the table of contents for the Appendix below:
\begin{enumerate}[label=\Alph*.]
    \item \hyperref[app:transition]{Subskill Transition Discovery}
        \begin{enumerate}[label=\Alph{enumi}.\arabic*.]
            \hypersetup{linkcolor=black}
            \item \hyperref[app:vis_transitions]{Visualization of Subskill Transitions}
            \item \hyperref[app:statistics_transitions]{Statistics of Subskill Transitions}
            \item \hyperref[app:change_point_detection]{Change-point Detection}
            \item \hyperref[app:soft_label]{Soft Labels for Subskill Transitions}
            \item \hyperref[app:validity_transition_labels]{Validity of Transition Labels with Demonstration Generator Labels}
        \end{enumerate}

    \item \hyperref[appendix:additional_results]{Additional Experiments}
        \begin{enumerate}[label=\Alph{enumi}.\arabic*.]
            \hypersetup{linkcolor=black}
            \item \hyperref[app:comparison_on_libero]{Comparison on LIBERO}
            \item \hyperref[app:scaling_longer_chunks]{Scaling to Longer Chunks}
            \item \hyperref[app:vlm_joint_training]{Results with VLM Joint Training}
            \item \hyperref[app:pred_exec_length]{Decoupling Prediction and Execution Chunk Lengths}
            \item \hyperref[app:generalization_to_other_vla]{Generalization to Other VLA Baselines}
            \item \hyperref[app:latency_overhead]{Latency Overhead Comparison}
            \item \hyperref[app:training_efficiency]{Training Efficiency Comparison}
            \item \hyperref[app:statistical_significance]{Statistical Significance}
        \end{enumerate}

    \item \hyperref[appendix:additional_ablation]{Additional Ablation Study}
        \begin{enumerate}[label=\Alph{enumi}.\arabic*.]
            \hypersetup{linkcolor=black}
            \item \hyperref[app:sensitivity_pelt]{Sensitivity to Change-point Detection Settings}
            \item \hyperref[app:effect_per_step_learnable_gate]{Effect of Per-step Learnable Gate}
            \item \hyperref[app:effect_soft_transition_label]{Effect of Soft Transition Label}
            \item \hyperref[app:effect_teacher_forcing_jittering]{Effect of Teacher Forcing and Jittering}
        \end{enumerate}

    \item \hyperref[appendix:additional_analysis]{Additional Analyses}
        \begin{enumerate}[label=\Alph{enumi}.\arabic*.]
            \hypersetup{linkcolor=black}
            \item \hyperref[app:accuracy_transition_timing_prediction]{Accuracy of Transition-timing Prediction}
            \item \hyperref[app:sensitivity_to_shift]{Sensitivity to Transition-timing Shift at Inference}
            \item \hyperref[app:prior_generated_timing]{Effect of the Prior on Generated Switching Timing}
            \item \hyperref[app:action_errors_dimension]{Decomposing Transition Errors by Action Component}
            \item \hyperref[app:idle_time]{Decomposing Idle Time in Real-robot Experiments}
        \end{enumerate}

    \item \hyperref[appendix:additional_details]{Additional Details}
        \begin{enumerate}[label=\Alph{enumi}.\arabic*.]
            \hypersetup{linkcolor=black}
            \item \hyperref[app:real-world]{Details on Real-robot Experiments}
            \item \hyperref[app:speedup_measure]{Details on Simulation Speedup Measurement}
            \item \hyperref[app:other_vlas]{Details on Other VLA Baselines}
            \item \hyperref[app:action_error_measure]{Details on Action Error Measurement}
            \item \hyperref[app:dataset]{Details on Dataset and Evaluation Protocols}
            \item \hyperref[app:pseudocode]{Pseudocode for Training and Inference}
        \end{enumerate}

    \item \hyperref[appendix:further_discussion]{Further Discussion}
        \begin{enumerate}[label=\Alph{enumi}.\arabic*.]
            \hypersetup{linkcolor=black}
            \item \hyperref[app:limitation]{Limitations}
            \item \hyperref[app:broader_impact]{Broader Impact}
        \end{enumerate}

\end{enumerate}

\clearpage

\vspace{-0.2cm}
\section{Subskill Transition Discovery}
\label{app:transition}
\vspace{-0.2cm}
To obtain subskill transition labels, we apply unsupervised subskill discovery~\citep{cotpc}
to the action sequence of each training demonstration,
which segments it into subskills without any manual annotation using change-point detection, \ie, PELT algorithm~\citep{pelt}.
Below, we first visualize examples of the detected transitions, report statistics of transitions, describe how they are obtained,
and then validate the transition labels.

\vspace{-0.2cm}
\subsection{Visualization of Subskill Transitions}
\label{app:vis_transitions}
\begin{figure}[h!]
    \centering
    \begin{subfigure}{0.85\textwidth}
        \centering
        \includegraphics[width=\linewidth]{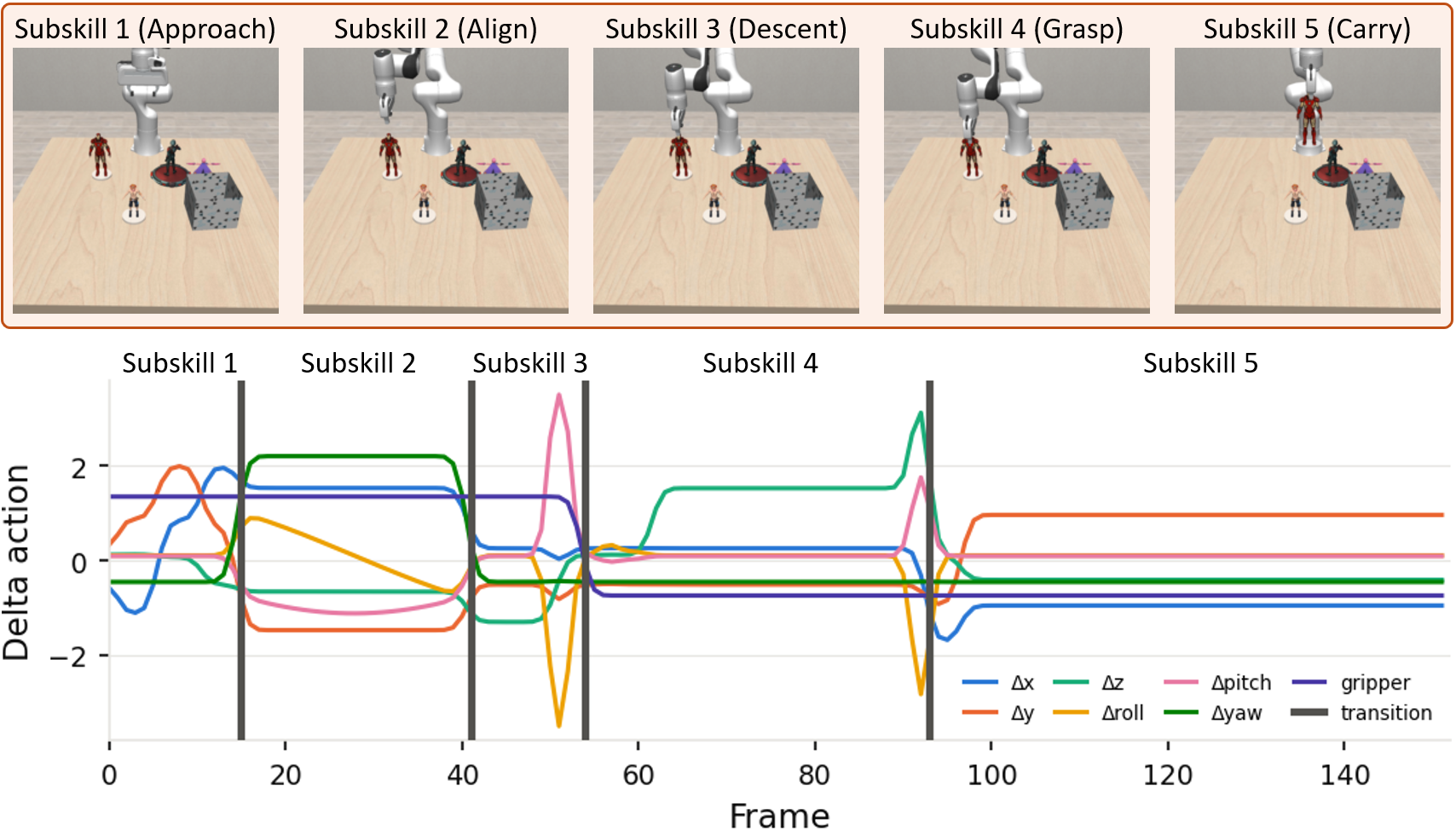}
        \vspace{-0.5cm}
        \caption{Language instruction: \textit{``Put the ironman into the giftbox.''}}
        \label{fig:transition_examples_a}
    \end{subfigure}\\[6pt]
    \begin{subfigure}{0.85\textwidth}
        \centering
        \includegraphics[width=\linewidth]{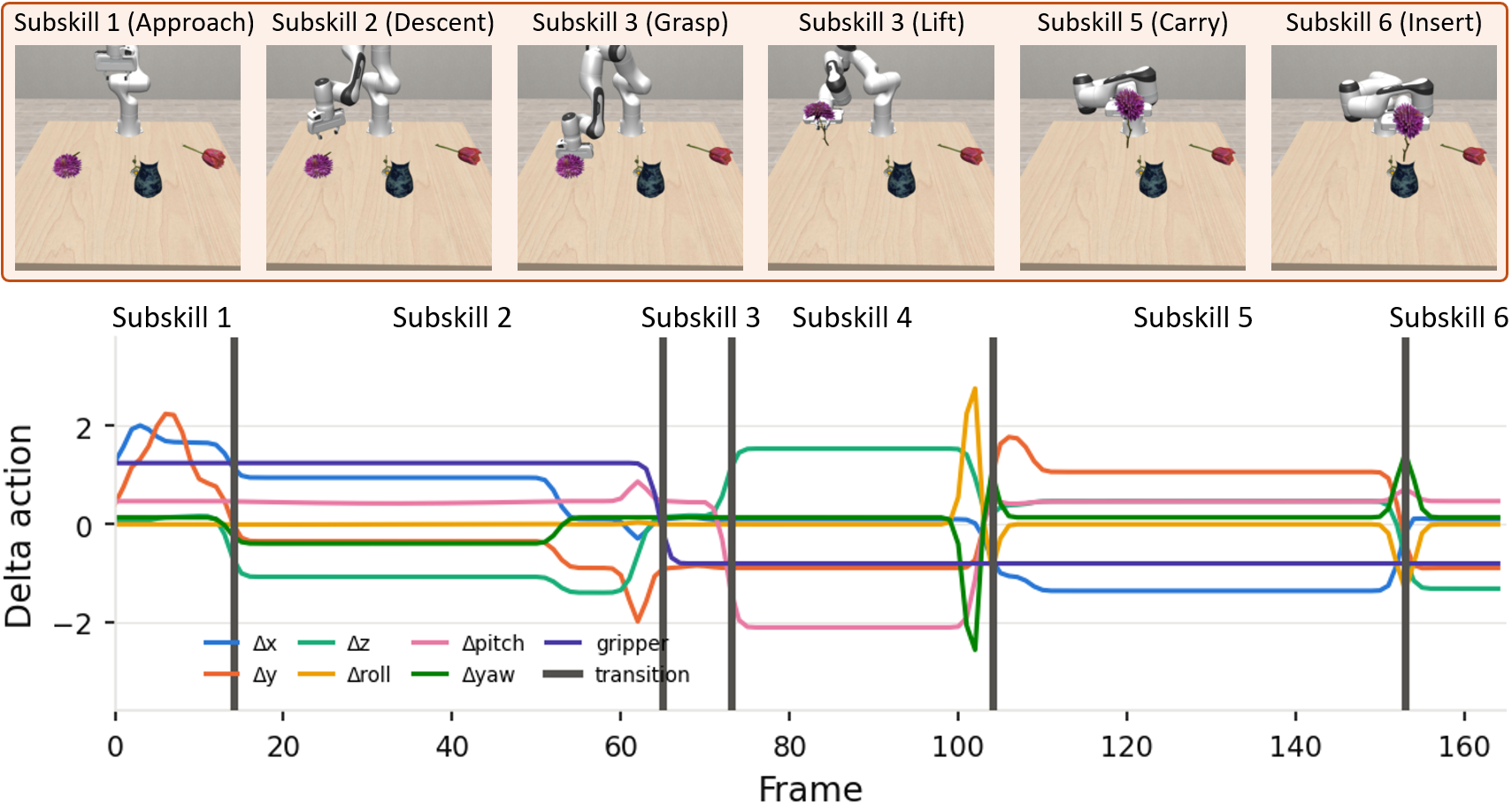}
        \vspace{-0.5cm}
        \caption{Language instruction: \textit{``Insert the chrysanthemum into the vase.''}}
        \label{fig:transition_examples_b}
    \end{subfigure}
    \vspace{-0.1cm}
    \caption{Visualization of subskill transitions.
        Change points detected by PELT~\citep{pelt} are overlaid on the action sequences of two VLABench demonstrations,
    alongside a representative frame for each subskill.
    Subskill names are generated with Claude Code~\citep{claudecode} for illustration only;
    only the transition points are used as labels for policy learning, not the semantics.
    }
    \vspace{-0.2cm}
    \label{fig:transition_examples}
\end{figure}

\vspace{-0.2cm}
Fig.~\ref{fig:transition_examples} shows the detected transitions on two VLABench demonstrations~\citep{vlabench},
overlaying the PELT change points on the action signal alongside representative frames of each subskill.
The subskill names shown in the figure are generated with Claude Code~\citep{claudecode} for illustration only;
RACE uses the transition points as labels for policy learning, not the subskill categories.
The change points fall at visible subskill boundaries, such as the switch from approach to grasp.

\vspace{-0.2cm}

\vspace{-0.2cm}
\subsection{Statistics of Subskill Transitions}
\label{app:statistics_transitions}
\vspace{-0.15cm}
Tab.~\ref{tab:transition_stats} summarizes the detected transitions on the VLABench training demonstrations.
Across the ten tasks, an episode contains $3.57$ change points on average,
ranging from $2.19$ for \textit{``select\_fruit''} to $5.59$ for \textit{``add\_condiment''}.
The number of transitions is not simply proportional to episode length:
\textit{``select\_fruit''} has among the longest episodes ($151$ steps) but the fewest transitions,
whereas \textit{``select\_chemistry\_tube''} has short episodes ($78$ steps) yet more transitions than several longer tasks.

\input{tab/transition_stats.tex}

\vspace{-0.2cm}
\subsection{Change-point Detection}
\label{app:change_point_detection}
\vspace{-0.15cm}
Following CoTPC~\citep{cotpc} and PRIMT~\citep{primt}, we segment each demonstration into subskills
with the PELT algorithm~\citep{pelt}.
Given a demonstration of $T$ actions $(\mathbf{a}_1,\ldots,\mathbf{a}_T)$,
where each $\mathbf{a}_t\in\mathbb{R}^{D}$ consists of a pose delta and a gripper command,
we form the input signal from the first difference of the pose delta (with angles wrapped to $[-\pi,\pi)$)
and the gripper command, and standardize each dimension within the episode,
$\tilde{\mathbf{a}}_t=(\mathbf{s}_t-\boldsymbol{\mu})/\boldsymbol{\sigma}$,
where $\mathbf{s}_t$ is the signal at step $t$,
$\boldsymbol{\mu},\boldsymbol{\sigma}\in\mathbb{R}^{D}$ are its per-dimension mean and standard deviation over the episode,
and the division is element-wise.
This normalization prevents action dimensions with different scales
(\eg, translation, rotation, and gripper control) from disproportionately influencing change-point detection.
PELT then identifies a set of change points $\mathcal{C}=\{c_1,\ldots,c_m\}$ by minimizing
\begin{align}
\sum_{i=0}^{m}
\operatorname{cost}\!\left(\tilde{\mathbf{a}}_{c_i:c_{i+1}}\right)
+ \beta m,
\qquad
c_0=1,\quad c_{m+1}=T+1,
\label{eq:pelt}
\end{align}
where $\tilde{\mathbf{a}}_{c_i:c_{i+1}}$ denotes the action segment between consecutive change points 
and $\beta$ controls the penalty for introducing additional change points.
We use a kernel-based segment cost with a Gaussian kernel
$k(\mathbf{u},\mathbf{v})=\exp(-\gamma\|\mathbf{u}-\mathbf{v}\|^2)$,
which captures changes in the distribution of actions across subskills,
such as a transition from steady translational motion to small corrective movements.
For a segment spanning steps $a$ through $b-1$, the cost is
\begin{align}
\operatorname{cost}\!\left(\tilde{\mathbf{a}}_{a:b}\right)
=
\sum_{t=a}^{b-1} k(\tilde{\mathbf{a}}_t,\tilde{\mathbf{a}}_t)
-
\frac{1}{b-a} \sum_{s,t=a}^{b-1} k(\tilde{\mathbf{a}}_s,\tilde{\mathbf{a}}_t).
\end{align}
Each change point $c$ is the first step of a new segment; 
we discard the terminal boundary.
We set $\beta=4$ for all benchmarks and set the bandwidth $\gamma$ per demonstration by the median heuristic,
\ie, the inverse of the median squared distance over all pairs of standardized action steps in that demonstration
($\gamma\approx0.1$ on our signals).

\vspace{-0.2cm}
\subsection{Soft Labels for Subskill Transitions}
\label{app:soft_label}
\vspace{-0.15cm}
PELT yields hard labels that mark a single step per transition.
Since a subskill transition may span several action steps, we convert these into soft labels.
For a chunk starting at time $t$, let $d_{t+h}$ be the number of steps from action step $t+h$
to its nearest change point in $\mathcal{C}$, and let $\mathbf{d}_t=(d_t,\ldots,d_{t+H-1})$.
The target $\mathbf{p}_t=(p_t,\ldots,p_{t+H-1})$ is
\begin{align}
    \mathbf{p}_t = \exp\!\left(-\frac{\mathbf{d}_t^{2}}{2\sigma_p^{2}}\right),
\end{align}
where the square and the exponential are applied element-wise, with $\sigma_p=1$.
Scores below $\exp(-R^2/2)$ for $R=3$ are set to zero, so each transition affects at most seven steps.
Steps beyond the end of an episode are clamped to its last action and receive a score of zero.
The target is computed once before training and stored alongside the demonstrations.

\begin{figure}[h]
    \centering
    \includegraphics[width=0.85\textwidth]{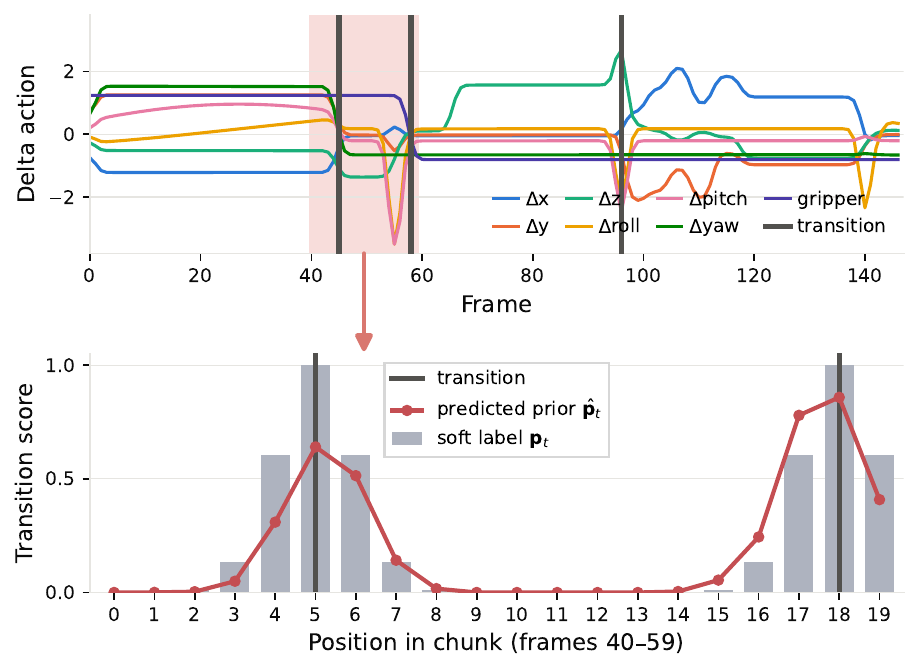}
    \vspace{-0.1cm}
    \caption{An example of transition-timing target and prediction on a held-out VLABench demonstration ($H_{\text{exec}}=20$).
    Top: standardized action signal and PELT change points; the pink-shaded chunk is enlarged in the bottom plot.
    Bottom: soft label $\mathbf{p}_t$ and predicted prior $\hat{\mathbf{p}}_t$ in this chunk.
    }
    \label{fig:prior_example}
    \vspace{-0.2cm}
\end{figure}

\vspace{-0.15cm}
\paragraph{Visualization of Soft Labels and Predicted Prior.}
Fig.~\ref{fig:prior_example} illustrates how this target is constructed and what the head predicts,
using a chunk of one of the held-out demonstrations (Appendix~\ref{app:validity_transition_labels}).
PELT detects change points from the standardized action signal (top),
here where a rotation ends and where the gripper closes.
Within the chunk, each change point is converted into a soft label that peaks at its position and spreads over neighboring steps (bottom, bars).
The head is trained to reproduce this soft label from the auxiliary pass,
so its prediction $\hat{\mathbf{p}}_t$ is a per-position score of how likely each action in the chunk is a transition,
rather than a single transition index.
In this example, the predicted prior peaks at both change points,
although it is lower and slightly wider than the label.
We chose this example to illustrate the target and prediction clearly;
the head does not capture every change point in general,
and its average accuracy is reported in Appendix~\ref{app:accuracy_transition_timing_prediction}.

\vspace{-0.2cm}
\subsection{Validity of Transition Labels with Demonstration Generator Labels}
\label{app:validity_transition_labels}
\vspace{-0.2cm}

To assess whether the PELT change points capture meaningful subskill transitions,
we compare them with reference boundaries obtained from the VLABench demonstration generator.
We conduct this analysis on 300 newly generated demonstrations spanning all ten VLABench tasks,
which are not used for training and serve as the held-out demonstrations throughout the appendix.
The reference boundaries are used only for evaluation.
We first describe how the generator provides the reference boundaries
and then evaluate how closely the PELT points align with them.

\vspace{-0.2cm}

\begin{figure}[h]
    \vspace{-0.2cm}
    \centering
    \includegraphics[width=0.9\textwidth]{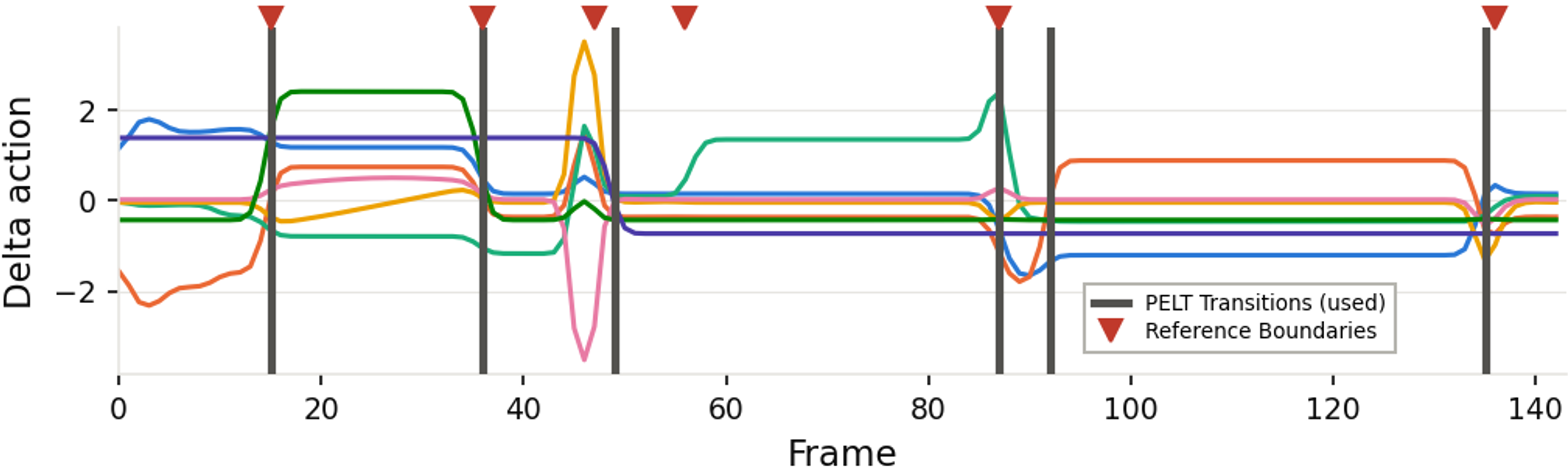}
    \vspace{-0.2cm}
    \caption{Qualitative comparison between PELT change points and reference subskill boundaries
    recorded by the VLABench demonstration generator.
    }
    \vspace{-0.3cm}
    \label{fig:transition_comparison}
\end{figure}

\paragraph{Reference Subskill Boundaries from the Demonstration Generator.}
The VLABench generator constructs each demonstration by executing a sequence of scripted skills.
These skills move the end effector through scene-dependent targets,
such as pre-grasp, grasp, and lift poses, interleaved with gripper operations.
We instrument the generator to record the steps at which it switches between these targets or primitives
and use them as reference subskill boundaries.
Intermediate motion-planning waypoints are excluded because they lie within a motion segment
and do not indicate a change in subskill.
This procedure yields an average of $4.5$ reference boundaries per episode.
Fig.~\ref{fig:transition_comparison} visualizes the resulting reference boundaries alongside the PELT change points.
We conduct this analysis only on VLABench,
whose scripted generator provides reference subskill boundaries unavailable in the other benchmarks,
\eg, human teleoperation used in LIBERO~\citep{libero}.

\vspace{-0.15cm}
\paragraph{Comparison with Reference Subskill Boundaries.}
Tab.~\ref{tab:label_validity} evaluates the three transition-label sources
considered in the ablation study of Tab.~\ref{tab:design_choices_labels}:
random labels, speed-minimum labels~\citep{pace}, and PELT change points.
We apply each source to 300 newly generated VLABench demonstrations
and compare its labels with the reference subskill boundaries recorded by the demonstration generator
($4.5$ per episode on average).
We use a tolerance of $\pm3$ steps,
the range over which each label is spread by the soft labels used for training (Appendix~\ref{app:soft_label}).
Precision is the fraction of labels that have a reference boundary within $\pm3$ steps,
and recall is the fraction of reference boundaries that have a label within $\pm3$ steps.
Both are counted per transition rather than per action step, without one-to-one matching,
and computed by pooling all episodes rather than averaging per episode.
Random labels draw the number of labels per episode uniformly from $\{1,\dots,9\}$
and place them at uniformly sampled steps.
PELT achieves substantially higher precision and recall than the alternative sources,
indicating closer alignment with the reference subskill boundaries.
This result is consistent with the transition-label ablation (Tab.~\ref{tab:design_choices_labels}),
where PELT outperforms speed-minimum labels.

\input{tab/label_validity.tex}

\vspace{-0.1cm}
\section{Additional Experiments}
\label{appendix:additional_results}

\vspace{-0.1cm}
\subsection{Comparison on LIBERO}
\label{app:comparison_on_libero}
\vspace{-0.1cm}
In Tab.~\ref{tab:libero_comparison}, we compare RACE on LIBERO with the $\pi_{0.5}$ baseline, its fine-tuned variants,
and other chunk-extension, adaptive chunking, and efficient VLA methods, all built on the same $\pi_{0.5}$ baseline.
The success rates of the $\pi_{0.5}$ baseline and Latent Bridge are taken from \citet{latentbridge},
and those of StreamPI~\citep{streampi}, ACoT-VLA~\citep{acotvla}, PolicyTrim~\citep{policytrim}, SAFE-Pruner~\citep{safepruner}, and GridS~\citep{grids} from their original papers,
while Spd is measured for all methods in our setup.
At every $H_{\text{exec}}$, RACE outperforms the fine-tuned $\pi_{0.5}$ executing the same number of actions.
With $H_{\text{exec}}\in\{10,15\}$, RACE performs comparably to recent state-of-the-art VLAs~\citep{streampi,acotvla},
which require more inference time than the $\pi_{0.5}$ baseline,
and outperforms adaptive chunking methods~\citep{autohorizon,pace}
and efficient VLA methods~\citep{latentbridge,safepruner,grids}
in both success rate and per-episode inference speedup.
Since these efficient VLA methods reduce the per-call cost, they are complementary to RACE and could be combined with it.
Even at $H_{\text{exec}}=20$, RACE remains competitive with the $\pi_{0.5}$ baseline
while requiring far less inference time per episode.
PolicyTrim~\citep{policytrim} achieves the largest speedup by optimizing task completion through reinforcement learning,
but its success rate is lower than that of RACE with $H_{\text{exec}}\in\{10,15\}$.
Moreover, its reinforcement learning incurs a substantial training cost,
as it collects fresh policy rollouts at every iteration and trains a separate model for each suite:
it takes 457 hours of wall-clock training per suite (about 1,828 hours for all four suites), whereas RACE takes 17 hours for all four suites,
both on four NVIDIA RTX A6000 GPUs (Appendix~\ref{app:training_efficiency}).

\input{tab/libero_comparison.tex}

\vspace{-0.1cm}
\subsection{Scaling to Longer Chunks}
\label{app:scaling_longer_chunks}
\vspace{-0.1cm}

In Tab.~\ref{tab:vlabench_long_chunks}, we evaluate RACE with execution chunk lengths ranging from
$H_{\text{exec}}=5$ to $40$ to examine the limits of chunk extension.
RACE maintains a higher success rate than the $\pi_{0.5}$ baseline
up to $H_{\text{exec}}=30$, but falls slightly below it at
$H_{\text{exec}}=40$.
These results show that RACE supports substantial chunk extension.
While longer chunks further reduce per-episode policy-inference time,
success rates decline at larger execution lengths.
This decline is likely because open-loop drift errors, which grow with the chunk length,
cannot be fully compensated by the transition-timing prior.

\input{tab/vlabench_long_chunks.tex}

\vspace{-0.1cm}
\subsection{Results with VLM Joint Training}
\label{app:vlm_joint_training}
\vspace{-0.1cm}

To examine whether the VLM frozen setting limits our framework,
we additionally update the VLM during post-training
and evaluate on VLABench with $H_{\text{exec}}=20$ in Tab.~\ref{tab:vlm_joint_training}.
VLM joint training improves the average success rate and progress score,
with the largest gains on the instruction and texture tracks that rely on the VLM.
Thus, RACE is not tied to the VLM frozen setting and further benefits from adapting the VLM.
We nevertheless keep the VLM frozen in the main experiments,
since joint training requires substantially more memory and compute (Tab.~\ref{tab:vlm_joint_training_cost}).

\input{tab/vlm_joint_training.tex}
\input{tab/vlm_joint_training_cost.tex}

\vspace{-0.1cm}
\subsection{Decoupling Prediction and Execution Chunk Lengths}
\label{app:pred_exec_length}
\vspace{-0.1cm}
On the real robot, each policy predicts twice as many actions as it executes ($H=2H_{\text{exec}}$),
since real-time chunking~\citep{rtc} generates the next chunk while the remaining actions are executed (Sec.~\ref{subsec:real_world}).
In the simulation experiments, in contrast, RACE and the fine-tuned $\pi_{0.5}$ predict and execute the same number of actions ($H=H_{\text{exec}}$)
to isolate the effect of extending the chunk,
since the policy then predicts exactly as far ahead as it executes.
We examine the decoupled setting of the real robot in simulation,
training both models to predict $H=40$ actions while executing only the first $H_{\text{exec}}=20$ on VLABench.
As shown in Tab.~\ref{tab:pred_exec_length}, predicting a longer chunk improves both models.
This is likely because errors grow toward the end of a chunk (as analyzed in Fig.~\ref{fig:drift_error} of the main manuscript),
so executing only its first half avoids the least accurate actions.

\input{tab/pred_exec_length.tex}

\vspace{-0.1cm}
\subsection{Generalization to Other VLA Baselines}
\label{app:generalization_to_other_vla}
\vspace{-0.1cm}
To examine the generality of our framework, we apply it to two other VLA baselines, 
GR00T-N1.6~\citep{gr00t} and OpenVLA-OFT~\citep{openvlaoft},
and evaluate them on LIBERO in Tab.~\ref{tab:backbones}.
GR00T-N1.6 generates actions by flow matching, like $\pi_{0.5}$, 
whereas OpenVLA-OFT produces actions through parallel decoding and regression.
The two baselines thus assess whether our transition-timing conditioning generalizes to another flow-matching model and beyond flow-based generation.
Implementation details for these VLA models are provided in Appendix~\ref{app:other_vlas}.
On GR00T-N1.6, RACE outperforms the fine-tuned variant at every chunk length,
and at $H_{\text{exec}}=16$ it even surpasses the baseline that executes half as many actions,
with the largest gains on the long-horizon suite at $H_{\text{exec}}\in\{16,24\}$, where fine-tuning degrades most.
On OpenVLA-OFT, RACE again outperforms the fine-tuned variant at every chunk length,
but neither recovers the baseline's success rate and the gain over fine-tuning is smaller.
One possible reason is the absence of a denoising process:
RACE is designed to condition the iterative refinement of the action expert (Fig.~\ref{fig:learned_gates}),
whereas OpenVLA-OFT regresses the chunk in a single pass.
\input{tab/libero_comparison_other_VLAs.tex}

\vspace{-0.1cm}
\subsection{Latency Overhead Comparison}
\label{app:latency_overhead}
\vspace{-0.1cm}
Tab.~\ref{tab:latency_comparison} compares the per-call latency overhead of RACE and ACoT-VLA~\citep{acotvla} over the $\pi_{0.5}$ baseline.
All models are measured on a single NVIDIA RTX A6000 GPU, using the official implementation codes for ACoT-VLA.
Since both methods share the same VLM and action expert,
the difference comes only from the modules each adds.
RACE adds the auxiliary one-step denoising pass and the transition-timing prediction head (Sec.~\ref{sec:head})
and the transition-conditioned modulation (Sec.~\ref{sec:cond}),
which together cost about $4\%$ of a policy call.
ACoT-VLA instead runs a second flow-matching denoising in its explicit action reasoner to produce a coarse reference trajectory,
and an additional cross-attention in its implicit action reasoner,
where the former alone costs about as much as the action expert itself.
This latency gap explains why ACoT-VLA achieves a lower speedup than RACE at the same $H_{\text{exec}}$ in Tab.~\ref{tab:vlabench}.
All latencies here are measured in the same benchmark setting for every model;
our current real-robot implementation of RACE runs without CUDA graphs and thus has a higher latency (Appendix~\ref{app:idle_time}).

\input{tab/latency_comparison.tex}

\vspace{-0.1cm}
\subsection{Training Efficiency Comparison}
\label{app:training_efficiency}
\vspace{-0.1cm}
To highlight the training efficiency of our framework, we compare the training cost of RACE with the fine-tuned $\pi_{0.5}$ and other methods,
such as ACoT-VLA~\citep{acotvla}, a recent state-of-the-art VLA, and PolicyTrim~\citep{policytrim}, a chunk-extension method,
on VLABench in Tab.~\ref{tab:training_efficiency_vlabench} and LIBERO in Tab.~\ref{tab:training_efficiency_libero}.
All methods are measured on four NVIDIA RTX A6000 GPUs, and the compared methods are run from their official implementations.
RACE adds only the transition-timing prediction head and the modulation parameters to the action expert,
so its training cost is only moderately higher than that of the fine-tuned $\pi_{0.5}$.
In contrast, ACoT-VLA also trains the visual encoder and its components with a larger batch,
taking more than an order of magnitude longer to train than RACE,
and PolicyTrim requires two-stage reinforcement learning with a much larger batch,
which incurs substantially higher training cost since every iteration collects fresh policy rollouts in the simulator.
Moreover, while the other methods train a single model on all four LIBERO suites,
PolicyTrim trains a separate model for each suite,
so covering all four suites takes an estimated 1,828 hours, over 100 times longer than RACE (Tab.~\ref{tab:training_efficiency_libero}).
RACE thus extends the action chunk with far less training cost than these methods.

\input{tab/training_efficiency_vlabench.tex}

\input{tab/training_efficiency_libero.tex}

\vspace{-0.1cm}
\subsection{Statistical Significance}
\label{app:statistical_significance}
\vspace{-0.1cm}
To examine how much the results vary across training runs,
we train RACE with $H_{\text{exec}}=20$ on VLABench with three different random seeds in Tab.~\ref{tab:seed_variance}.
The average success rate is $33.3\pm0.5$ (mean $\pm$ standard deviation),
and every run outperforms the fine-tuned $\pi_{0.5}$ in Tab.~\ref{tab:vlabench}, which is trained once,
by a margin much larger than this variation.
Individual tracks vary more across runs, so small differences on a single track should be interpreted with care.
The model in Tab.~\ref{tab:vlabench} is the first run, whose results are close to the mean.

\input{tab/seed_variance.tex}

\vspace{-0.1cm}
\section{Additional Ablation Study}
\label{appendix:additional_ablation}
\vspace{-0.1cm}

\vspace{-0.1cm}
\subsection{Sensitivity to Change-point Detection Settings}
\label{app:sensitivity_pelt}
\vspace{-0.1cm}
To examine the sensitivity of RACE to the PELT settings, we vary the penalty $\beta$ in Eq.~(\ref{eq:pelt}) of Appendix~\ref{app:change_point_detection},
which controls the number of detected transitions,
and retrain our framework with the resulting labels on VLABench with $H_{\text{exec}}=20$ in Tab.~\ref{tab:sensitivity_pelt}.
Performance stays stable from $\beta=2$ to $8$ despite a threefold change in label density,
so RACE is not sensitive to the PELT settings.

\input{tab/sensitivity_pelt.tex}

\vspace{-0.1cm}
\subsection{Effect of Per-step Learnable Gate}
\label{app:effect_per_step_learnable_gate}
\vspace{-0.1cm}
Tab.~\ref{tab:gate_ablation} evaluates the learnable per-step gate
$\alpha^{k}$ in Eq.~(\ref{eq:race_denoising}), which controls the strength
of prior conditioning at each denoising step.
Replacing these gates with a fixed value of $\alpha^{k}=1$ applies the prior
equally at every step and performs slightly worse.
This suggests that letting the model adjust the conditioning strength per step is beneficial,
in line with the learned gates in Fig.~\ref{fig:learned_gates},
which take on a step-dependent pattern after training.

\input{tab/gate_ablation.tex}

\vspace{-0.1cm}
\subsection{Effect of Soft Transition Label}
\label{app:effect_soft_transition_label}
\vspace{-0.1cm}
As described in Sec.~\ref{sec:head} and Appendix~\ref{app:soft_label},
we convert the hard transition labels from PELT into soft labels.
Tab.~\ref{tab:soft_label} compares these two labels (hard labels and soft labels) on VLABench with $H_{\text{exec}}=20$.
We observe that training with hard labels (\ie, one-hot labels at every change point) lowers the success rate,
suggesting that spreading each transition over its neighboring steps gives a more useful target than a single step,
as a subskill transition spans several actions and the detected points may not exactly match the true transition steps.

\input{tab/soft_label.tex}

\vspace{-0.1cm}
\subsection{Effect of Teacher Forcing and Jittering}
\label{app:effect_teacher_forcing_jittering}
\vspace{-0.1cm}

During training, we condition the action expert on the transition target $\mathbf{p}_t$ instead of the predicted prior $\hat{\mathbf{p}}_t$,
\ie, teacher forcing,
avoiding reliance on inaccurate timing predictions early in training.
To improve robustness to imperfect priors,
we randomly shift the target by one action step with probability $0.5$ when using it for conditioning, \ie, jittering,
while keeping the target for $\mathcal{L}_{\mathrm{timing}}$ unchanged.
At inference, we condition on the predicted prior $\hat{\mathbf{p}}_t$.
Tab.~\ref{tab:design_choices_training} evaluates both choices: each has only a marginal effect on VLABench,
indicating that RACE is not sensitive to them.
However, jittering mitigates the performance drop under inference-time shifts of several steps
(Tab.~\ref{tab:sensitivity_shift} in Appendix~\ref{app:sensitivity_to_shift}).

\input{tab/teacher_forcing_jittering.tex}

\vspace{-0.1cm}
\section{Additional Analyses}
\label{appendix:additional_analysis}
\vspace{-0.1cm}

\vspace{-0.1cm}
\subsection{Accuracy of Transition-timing Prediction}
\label{app:accuracy_transition_timing_prediction}
\vspace{-0.1cm}
To check how accurately the head predicts transition timing,
we compare its predictions with the PELT labels in Tab.~\ref{tab:timing_accuracy}
on 5,000 VLABench training demonstrations and the 300 held-out demonstrations (Appendix~\ref{app:validity_transition_labels}).
All thresholds are selected on the training demonstrations and applied unchanged to the held-out demonstrations.
These results measure agreement with the PELT labels on demonstration states,
not on the states visited during policy rollouts.

\vspace{-0.15cm}
\paragraph{Evaluation Metrics.}
We evaluate three aspects:
\textbf{Transition-chunk detection} checks whether a chunk contains a transition,
scoring each chunk by the maximum of the predicted prior $\hat{\mathbf{p}}_t$ with a threshold of $0.14$.
\textbf{Transition-point localization} checks where the transition occurs:
in each chunk containing a transition, we check whether the position with the highest prior
lies within $\pm1$ or $\pm3$ steps of the nearest PELT change point.
\textbf{Multi-transition localization} checks how many transitions in a chunk are captured,
since a chunk may contain several transitions while localization credits only the highest peak.
We take every peak above a threshold ($0.35$, $0.35$, and $0.40$ for $H_{\text{exec}}=10$, $15$, and $20$) as a predicted transition
and match it to the change points within $\pm1$ or $\pm3$ steps.
Precision is the fraction of predicted transitions that are matched, and recall is the fraction of change points that are matched.
This threshold is higher than that for detection because spurious peaks count as false positives here.
Since localization always uses a single peak per chunk, its values are not directly comparable to recall.
For transition labels and predictions, we use tolerances of $\pm1$ and $\pm3$ steps,
matching the resolution at which RACE is trained:
the soft labels have a width of $\sigma_p=1$ and are truncated at $R=3$ steps (Appendix~\ref{app:soft_label}),
and the conditioning target is jittered by one step during training (Appendix~\ref{app:effect_teacher_forcing_jittering}).

\input{tab/timing_accuracy.tex}

\vspace{-0.15cm}
\paragraph{Accuracy Results of Transition-timing Head.}
On training demonstrations, detection and localization are well above random guessing
(50\% for detection; 28\%, 19\%, and 15\% within $\pm1$ step for localization at $H_{\text{exec}}=10$, $15$, and $20$),
and most predicted peaks fall within the range covered by the soft labels (Appendix~\ref{app:soft_label}).
On held-out demonstrations, accuracy drops, most for localization with longer chunks,
but remains well above random guessing.
We compare detection between the two sets by AUROC,
since accuracy depends on the fraction of transition chunks, which is higher on held-out demonstrations.
For multi-transition localization, precision is high, so most predicted transitions are real,
but recall is lower: of $3.6$ and $3.8$ transitions per episode on training and held-out demonstrations,
about one and up to $1.6$ are missed, respectively.
The slightly higher held-out precision within $\pm3$ steps comes from fewer predicted peaks, not better prediction.
Overall, the head predicts transitions with high precision but misses some of them.
Although this evaluation is limited to demonstration states,
the prior also affects the generated actions and rollouts:
RACE switches the gripper within one step of the correct step more often (Tab.~\ref{tab:gripper_timing}),
and shifting the prior by one step lowers the success rate (Appendix~\ref{app:sensitivity_to_shift}).

\vspace{-0.1cm}
\subsection{Sensitivity to Transition-timing Shift at Inference}
\label{app:sensitivity_to_shift}
\vspace{-0.1cm}

To examine the sensitivity to transition-timing shift,
we shift the predicted transition timing by $\delta$ steps at inference and report 
success rates on VLABench with $H_{\text{exec}}=20$ in Tab.~\ref{tab:sensitivity_shift};
peaks shifted beyond the chunk are dropped, so large shifts also remove part of the prior.
Any shift lowers performance, even by a single step,
showing that the model uses the predicted prior rather than ignoring it.
Without jittering (Appendix~\ref{app:effect_teacher_forcing_jittering}),
the model degrades more for shifts of several steps (\eg, $\pm3$ steps),
while the two models perform similarly for single-step shifts.
Jittering thus mitigates large drops under timing errors of several steps.
The partial recovery at the largest shifts reflects the removed prior rather than tolerance to the shift,
reaching a level similar to the zero-prior setting of Tab.~\ref{tab:inference_prior_perturbing_replacement_wrap} in the main manuscript.

\input{tab/sensitivity_shift.tex}

\vspace{-0.1cm}
\subsection{Effect of the Prior on Generated Switching Timing}
\label{app:prior_generated_timing}
\vspace{-0.1cm}
In Tab.~\ref{tab:prior_intervention_gripper}, to see how the prior affects the generated actions,
we shift the predicted prior by $k$ steps or replace it with zeros at inference,
while keeping the observation, noise, and weights fixed,
and measure when the generated gripper command switches on VLABench with $H_{\text{exec}}=20$.
We use the gripper since its binary command switches at a well-defined step, unlike translation and rotation.
Shifting the prior moves the generated switch in the same direction,
at 0.23 and 0.28 steps per step of shift on training and held-out demonstrations
(95\% bootstrap confidence intervals exclude zero for every $k$),
showing that the prior directly controls when the generated actions switch.
Since the switch moves by only about a quarter of the shift,
the model also relies on features learned from the observation.
Replacing the prior with zeros leaves the switch offset nearly unchanged but increases missed switches,
consistent with the lower success rate of the zero-prior setting in Tab.~\ref{tab:inference_prior_perturbing_replacement_wrap}.

\input{tab/prior_intervention_gripper.tex}

\vspace{-0.1cm}
\subsection{Decomposing Transition Errors by Action Component}
\label{app:action_errors_dimension}
\vspace{-0.1cm}
The action error in Fig.~\ref{fig:drift_error} averages all seven action dimensions,
which can hide differences between action components.
We therefore split the errors at transitions into translation ($x$, $y$, $z$), rotation (roll, pitch, yaw), and gripper components,
using 500 VLABench training demonstrations at $H_{\text{exec}}=20$.
Errors are measured as in Appendix~\ref{app:action_error_measure} and averaged over four noise seeds,
except that transitions here include the chunk positions within $\pm1$ step of a PELT change point.

As shown in Tab.~\ref{tab:transition_error_by_component},
translation has the largest errors and thus dominates the overall error.
Separating the components shows that RACE reduces errors in all three components over the fine-tuned $\pi_{0.5}$,
with larger reductions in rotation and gripper.
Since we do not subtract open-loop drift here and RACE also reduces errors away from transitions,
these results show that errors at transitions decrease, but not that they decrease more than elsewhere.
\input{tab/transition_error_by_component.tex}

\vspace{-0.15cm}
\paragraph{Gripper Switching Timing.}
Since the gripper command takes only two values, its error mainly reflects when the gripper opens or closes.
We measure this timing directly in Tab.~\ref{tab:gripper_timing}
on chunks where the demonstrated gripper command switches exactly once,
taking the predicted switch as the step where the predicted gripper value changes sign.
Compared with the fine-tuned $\pi_{0.5}$, RACE switches within $\pm1$ and $\pm3$ steps of the correct step more often and misses fewer switches
on both training and held-out demonstrations, with a similar false-switch rate,
suggesting that the gain is not due to predicting more switches.
RACE also switches at the exact step more often on training demonstrations,
while this gain is not significant on held-out demonstrations.
Mean error is also lower on training demonstrations, and lower but not significantly so on held-out demonstrations.
\input{tab/gripper_timing.tex}

\vspace{-0.1cm}
\subsection{Decomposing Idle Time in Real-robot Experiments}
\label{app:idle_time}
\vspace{-0.1cm}

Tab.~\ref{tab:idle_decomposition} decomposes the idle time in Tab.~\ref{tab:realworld_pnp} for real-robot experiments,
which measures the total time the robot receives no new action command during an episode.

\vspace{-0.15cm}
\paragraph{When the robot pauses.}
With asynchronous execution, each policy call proceeds as follows:
the client captures new frames from both cameras, sends the observation to the server,
and receives the next chunk, while the robot keeps executing the remaining actions of the current chunk.
Since the robot keeps moving during inference, real-time chunking skips the first actions of the new chunk,
which correspond to the time elapsed during inference, and executes the rest.
We call the time to execute these remaining actions the \textit{execution buffer}:
the number of actions left in a chunk when it arrives, divided by the control frequency of 30\,Hz.
The robot pauses only when the latency of a policy call exceeds this buffer.

\input{tab/idle_decomposition.tex}

\vspace{-0.15cm}
\paragraph{Breakdown of the Idle Time.}
With $H_{\text{exec}}=5$, about three of the ten actions in a chunk are skipped on arrival,
leaving an execution buffer of about seven actions,
which is slightly shorter than the latency per call on average.
The robot therefore executes the whole chunk before the next one arrives,
about seven actions per call rather than five,
and pauses at most policy calls; these pauses account for most of the idle time.
The measured pause per call is slightly larger than this average margin,
because calls that finish within the buffer contribute zero rather than a negative pause.
With $H_{\text{exec}}=20$, the buffer is several times longer than the latency,
so no pause occurs after the first policy call.
Since the pauses depend on the per-call latency, a large part of which is camera capture in our setup,
the chunk length needed to remove them depends on the deployment setup.
The remaining idle time comes from two fixed costs that do not depend on the chunk length:
the startup, where the robot waits for the first camera capture and the first policy call,
and the control-loop overhead, such as timer jitter.
RACE has a longer server latency than the fine-tuned $\pi_{0.5}$ because our real-time chunking implementation
runs it without CUDA graphs, as its modulation changes at every denoising step.
It thus skips more actions on arrival, which shortens its execution buffer.
This stems from our implementation rather than from RACE itself,
and a CUDA-graph-compatible implementation could reduce this extra latency.
Even so, the longer execution buffer hides this latency, yielding the same idle time.

\vspace{-0.1cm}
\section{Additional Details}
\label{appendix:additional_details}

\vspace{-0.1cm}
\subsection{Details on Real-robot Experiments}
\label{app:real-world}
\vspace{-0.1cm}
In this section, we describe the task and training demonstrations, robot setup, evaluation metrics,
and implementation details for the real-robot results in Tab.~\ref{tab:realworld_pnp}.

\begin{figure}[h]
    \centering
    \begin{subfigure}[t]{0.19\textwidth}
        \centering
        \includegraphics[width=\linewidth]{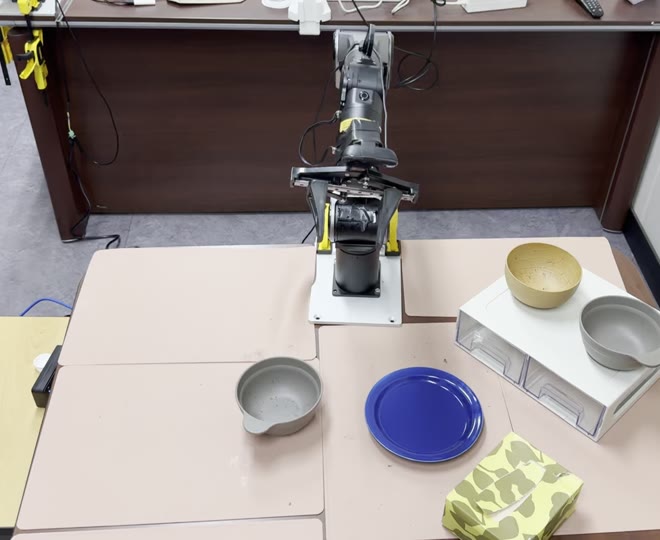}
        \caption{Initial state}
    \end{subfigure}\hfill
    \begin{subfigure}[t]{0.19\textwidth}
        \centering
        \includegraphics[width=\linewidth]{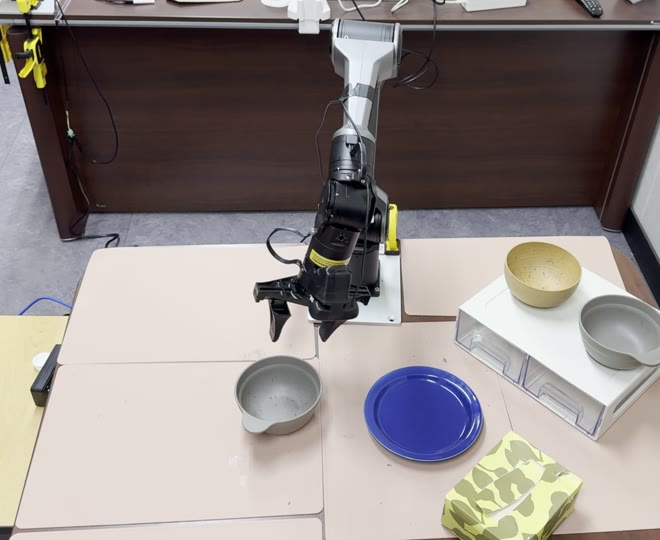}
        \caption{Approach}
    \end{subfigure}\hfill
    \begin{subfigure}[t]{0.19\textwidth}
        \centering
        \includegraphics[width=\linewidth]{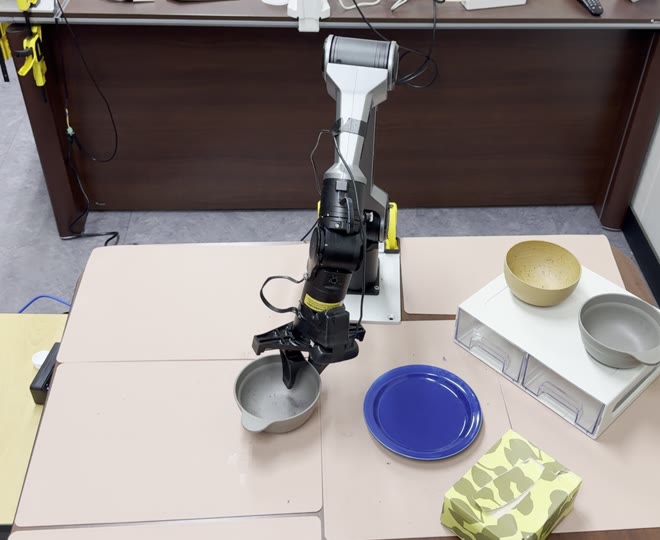}
        \caption{Grasp}
    \end{subfigure}\hfill
    \begin{subfigure}[t]{0.19\textwidth}
        \centering
        \includegraphics[width=\linewidth]{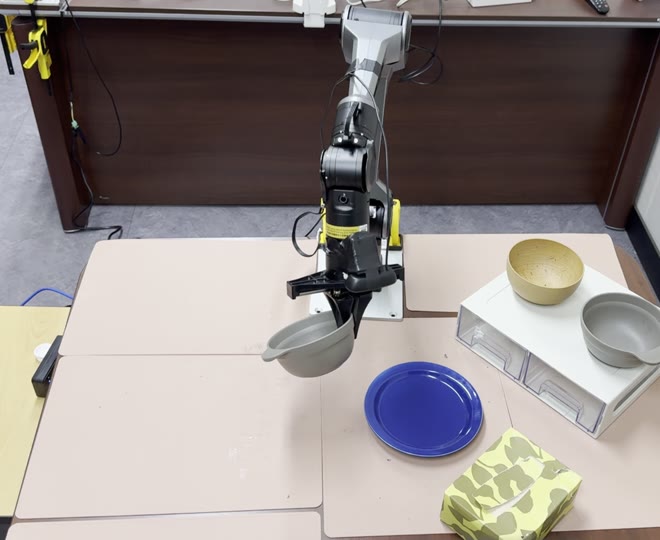}
        \caption{Lift}
    \end{subfigure}\hfill
    \begin{subfigure}[t]{0.19\textwidth}
        \centering
        \includegraphics[width=\linewidth]{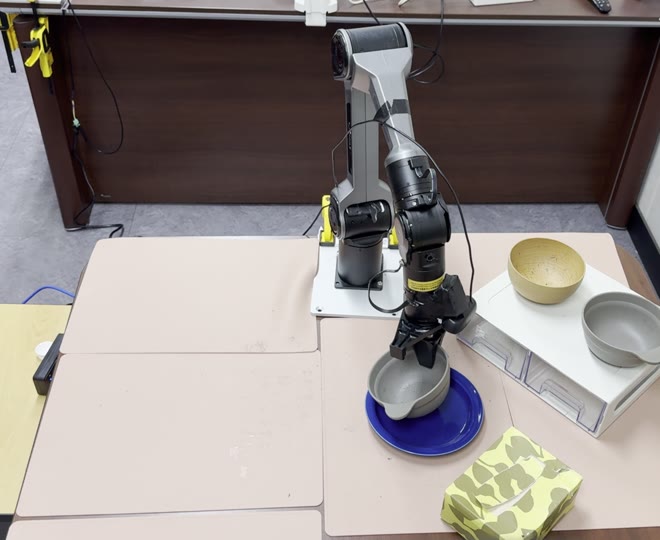}
        \caption{Place}
    \end{subfigure}
    \vspace{-0.2cm}
    \caption{Task Visualization.
    Keyframes of a successful rollout of RACE from the front view.}
    \label{fig:realworld_task}
\end{figure}

\vspace{-0.15cm}
\paragraph{Task Description.}
We evaluate a pick-and-place task with the instruction
\textit{``Pick up the gray bowl next to the plate and place it on the plate.''}
A trial is successful if the robot grasps the bowl and places it stably on the plate.
Across trials, we vary the initial positions of the bowl and the plate
and use the same set of initial placements for all methods.
For training, we use 30 demonstrations~\citep{selfcompensatingvla} of this task collected on the robotic platform,
each recording images from both cameras, the robot state, and the action at 30~Hz,
where the action consists of the absolute positions of the six joints and the gripper.
Fig.~\ref{fig:realworld_task} visualizes the task with a successful rollout of RACE at $H_{\text{exec}}=20$.

\vspace{-0.1cm}

\vspace{-0.15cm}
\paragraph{Robot Setup.}
We use an AgileX PiPER robotic arm, a 6-DoF lightweight arm with a 626-mm reach and a 1.5-kg payload, equipped with a gripper.
A wrist camera and a front camera provide the visual observations.

\vspace{-0.15cm}
\paragraph{Evaluation Metric.}
We run 50 trials per method and report the success rate, task-completion time, and idle time.
Task-completion time is measured from the start of a trial until the task is completed,
and idle time is the total time the robot receives no new action command during an episode
(decomposed in Appendix~\ref{app:idle_time}).
Both times are averaged over successful trials.
All methods use asynchronous execution with real-time chunking~\citep{rtc}
and predict a chunk twice as long as they execute ($H=2H_{\text{exec}}$).

\vspace{-0.15cm}
\paragraph{Implementation Details.}
All policies are fine-tuned from the pre-trained $\pi_{0.5}$ base model on the 30 demonstrations for 20K steps.
We train the fine-tuned $\pi_{0.5}$ with chunk lengths $H=10$ and $H=40$, and RACE with $H=40$.
We use AdamW with gradient clipping at 1.0, a batch size of 32,
and a cosine learning-rate schedule with 1K warmup steps, decaying from $2.5\times10^{-5}$ to $2.5\times10^{-6}$, without EMA.
Actions are predicted as joint displacements from the first state of each chunk (the gripper command is absolute)
and converted back to absolute joint targets before execution.
For RACE, we use the same transition-label and loss settings as in simulation.
At deployment, policies run asynchronously and request the next chunk after executing $H_{\text{exec}}=H/2$ actions,
\eg, a policy with $H=40$ requests the next chunk after executing 20 actions,
while the remaining 20 actions cover the inference time.
On the real robot, $H_{\text{exec}}$ thus denotes the configured replanning interval.
For $H=10$, the latency exceeds the execution time of the remaining actions,
so the robot executes the whole chunk before the next one arrives,
and each call executes about seven actions on average rather than five (Appendix~\ref{app:idle_time}).

\vspace{-0.1cm}
\subsection{Details on Simulation Speedup Measurement}
\label{app:speedup_measure}

\vspace{-0.15cm}
\paragraph{Simulation Experiments.}
We report the per-episode inference speedup (Spd$\uparrow$) in the main comparisons
(Tab.~\ref{tab:vlabench}, Tab.~\ref{tab:robocasa_h50}, and Tab.~\ref{tab:libero_comparison}).
For each model, we measure the latency of a single policy call
and the average number of policy calls per episode
over the episodes that both the model and the $\pi_{0.5}$ baseline complete successfully;
the speedup is the ratio of this product for the $\pi_{0.5}$ baseline to that for the model.
Formally, for the model $\theta$,
\begin{align}
    \mathrm{Spd}_{\theta}
    =
    \frac{\ell_{\pi_{0.5}}\,\bar{N}_{\pi_{0.5}}}
         {\ell_{\theta}\,\bar{N}_{\theta}},
    \label{eq:speedup}
\end{align}
where $\ell_{\theta}$ is the latency of a single policy call
and $\bar{N}_{\theta}$ is the average number of policy calls per episode over these common successful episodes,
where an episode of $T$ steps requires $\lceil T/H_{\text{exec}}\rceil$ policy calls.
Similar to PolicyTrim~\citep{policytrim}, this measure counts only the time spent waiting for the policy
and excludes simulator execution time, which depends on the simulator rather than on the policy
and does not exist on a real robot.
Unlike PolicyTrim, which compares only the number of policy calls,
we also account for the per-call latency, since RACE adds a small inference overhead.
Spd is thus largely determined by $H_{\text{exec}}$ and the per-call latency.
With synchronous execution, this reduction directly shortens the pauses between policy calls;
with asynchronous execution, longer chunks instead allow the remaining latency to be hidden (Appendix~\ref{app:idle_time}).
Beyond deployment, fewer policy calls also reduce the inference cost of large-scale simulation rollouts,
such as policy evaluation or data collection for reinforcement learning,
although the overall gain there depends on the simulator step time excluded from Spd.

\vspace{-0.15cm}
\paragraph{Breakdown and Robustness.}
Since Spd compares the policy-inference time per episode,
a higher value could also come from shorter episodes rather than fewer policy calls.
To separate these factors, Tab.~\ref{tab:speedup_breakdown} breaks down the speedup at $H_{\text{exec}}=20$ on VLABench
over the 127 episodes that all three models complete successfully,
which cover 32 of the 50 task--track pairs (59, 22, 19, 8, and 19 episodes from the five tracks).
The three models take a similar number of environment steps per episode,
so the speedup comes from fewer policy calls rather than shorter episodes.
RACE makes slightly more calls than the fine-tuned $\pi_{0.5}$ and adds a small per-call overhead
(Appendix~\ref{app:latency_overhead}), so its speedup is slightly lower.
In addition, since Spd is computed only over successful episodes,
it could change with which episodes are included.
Tab.~\ref{tab:speedup_sets} shows that the speedup remains similar across different sets of episodes,
including all episodes regardless of success.
Spd should thus be read together with the success rate, where RACE outperforms the fine-tuned $\pi_{0.5}$ (Tab.~\ref{tab:vlabench}).
\input{tab/speedup_breakdown.tex}
\input{tab/speedup_sets.tex}

\vspace{-0.15cm}
\paragraph{Real-robot Experiments.}
On real robots, we instead report task-completion time and idle time (Sec.~\ref{subsec:real_world}).
Task-completion time is measured from the start of a trial until the task is completed,
and idle time is the total time the robot receives no new action command during an episode
(decomposed in Appendix~\ref{app:idle_time}).

\vspace{-0.1cm}
\subsection{Details on Other VLA Baselines}
\label{app:other_vlas}
\vspace{-0.1cm}
We describe how RACE is applied to the two backbones in Tab.~\ref{tab:backbones}.
\textbf{1) GR00T-N1.6:}
like $\pi_{0.5}$, it generates actions with a flow-matching action head conditioned on VLM features,
so RACE transfers with a single change in how the prior is injected.
We use an auxiliary one-step denoising pass to obtain action features
and predict the transition-timing prior together with the cached VLM features.
Unlike the action expert of $\pi_{0.5}$, which uses adaptive RMSNorm,
the action head of GR00T-N1.6 is a diffusion transformer with adaptive LayerNorm (AdaLN)~\citep{dit} conditioned on the denoising timestep.
We therefore inject the prior into every subsequent denoising step through these AdaLN layers,
modulating their scale and shift token-wise as in Eq.~\ref{eq:adarms}.
\textbf{2) OpenVLA-OFT:} it regresses the chunk with a single language model and has no denoising process.
We first forward only the image and instruction tokens and cache their hidden states,
and then replace the auxiliary denoising pass by an unmodulated forward pass of the action placeholder tokens
to obtain the action features for the prediction head,
and finally run them a second time with the prior injected at every RMSNorm layer of the action positions.
Since there is no denoising loop, there is a single modulated pass and no per-step gate.

\vspace{-0.1cm}
\subsection{Details on Action Error Measurement}
\label{app:action_error_measure}
\vspace{-0.1cm}
We measure action errors on VLABench demonstrations,
since ground-truth actions are available only for demonstration states and not for the states visited during rollouts.
At each policy call, the policy predicts a chunk from the current observation,
and we compare the final output of the denoising process with the ground-truth actions of the demonstration over the same steps.
The action error of each action is the mean squared error (MSE) between the predicted and ground-truth actions
in the normalized action space, averaged over all seven action dimensions.
Since transition steps also involve larger changes in action,
comparing errors at transition and non-transition actions indicates where errors concentrate
rather than separating the causes of execution failures.

\vspace{-0.15cm}
\paragraph{Offset from transition point.}
For Fig.~\ref{fig:motivation}, we align the errors at transitions.
For each PELT change point at position $c$ within a chunk,
we take the error of the action at position $c+\delta$ for offsets $\delta\in\{-8,\ldots,8\}$,
keeping only offsets that lie within the chunk,
and average it over all change points.
A chunk containing multiple change points contributes each of them separately.
An offset of $\delta=0$ corresponds to the transition point.

\vspace{-0.15cm}
\paragraph{Position within chunk.}
For Fig.~\ref{fig:drift_error}, we group the actions by their position within the chunk.
At each position, we average the error of actions that coincide with a change point (\ie, transition error)
and that of the remaining actions (\ie, non-transition error).

\vspace{-0.15cm}
\paragraph{Matching local action variation.}
Because PELT detects changes in action dynamics,
larger errors at transitions could reflect larger local action variations.
To examine this, we compare the error at each transition with those at non-transition positions,
at least three steps from any change point,
matched by position within the chunk and by decile of the second-order action difference
$c_h=\|\mathbf{a}_h-2\mathbf{a}_{h-1}+\mathbf{a}_{h-2}\|_2$ of the demonstrated actions,
on 500 VLABench training demonstrations with $H_{\text{exec}}=20$ for $\pi_{0.5}$ and RACE as in Fig.~\ref{fig:motivation} (Tab.~\ref{tab:matched_action_change}).
Transition errors remain about 1.3 times those at the matched positions for both policies,
so the error gap persists after matching on chunk position and this measure of local action variation.

\input{tab/matched_action_change.tex}

\vspace{-0.1cm}
\subsection{Details on Dataset and Evaluation Protocols}
\label{app:dataset}

\vspace{-0.15cm}
\paragraph{VLABench.}
VLABench~\citep{vlabench} is a MuJoCo-based benchmark for long-horizon,
language-conditioned manipulation.
It contains ten tasks with 500 scripted demonstrations per task,
for a total of 5,000 training demonstrations.
Each task is evaluated under five tracks:
\textit{in-distribution}, which follows the training distribution;
\textit{unseen categories}, which changes the object categories;
\textit{commonsense}, which requires reasoning beyond explicit instructions;
\textit{semantic instructions}, which varies the linguistic specification; and
\textit{unseen textures}, which changes object appearance.
For each track, we run 50 episodes per task, resulting in 500 episodes per track
and 2,500 evaluation episodes in total.
We report success rate (SR) for full task completion and progress score (PS) for partial completion.
We first average results across the ten tasks within each track and then report
the mean across the five tracks.

\vspace{-0.15cm}
\paragraph{RoboCasa-H50.}
RoboCasa~\citep{robocasa} provides visually diverse kitchen environments with
randomized layouts, fixtures, objects, and object placements.
We use the RoboCasa-H50 setting, which contains 50 human-teleoperated demonstrations
for each of 25 kitchen-manipulation tasks, totaling 1,250 training demonstrations.
The tasks cover eight categories:
pick-and-place, doors, drawers, sinks, stoves, coffee machines, microwaves, and navigation.
We evaluate each task over 50 episodes, yielding 1,250 evaluation episodes per model.
Success is determined by the task-specific completion predicate.
We average success rates within each category and compute the overall score
weighted by the number of tasks per category.

\vspace{-0.15cm}
\paragraph{LIBERO.}
We evaluate on the four standard LIBERO suites~\citep{libero}:
Spatial, Object, Goal, and Long.
Each suite contains ten tasks with 50 human-teleoperated demonstrations per task.
Thus, we train on 2,000 demonstrations across the 40 tasks.
For evaluation, we run 50 episodes per task using the benchmark-provided initial states,
resulting in 500 episodes per suite and 2,000 episodes per model.
A rollout is successful when the task-specific goal is achieved within the episode horizon.
We report the mean success rate for each suite and the average across all four suites.

\vspace{-0.1cm}
\subsection{Pseudocode for Training and Inference}
\label{app:pseudocode}
\vspace{-0.1cm}
In this section, to provide the exact training and inference procedures of RACE,
we present their pseudocode in Alg.~\ref{alg:train} and Alg.~\ref{alg:infer}.
Unlike Sec.~\ref{sec:overview}, which omits it for brevity, we write the flow time $s\in[0,1]$ explicitly;
it runs from the noise ($s=0$) to the action chunk ($s=1$).

\vspace{-0.15cm}
\paragraph{Training Procedure.}
Each training step first runs the auxiliary pass on the noise sample without modulation,
from which the prediction head estimates the prior and $\mathcal{L}_{\mathrm{timing}}$ and $\mathcal{L}_{\mathrm{aux}}$ are computed.
The full pass then samples the flow time $s$ continuously as in $\pi_{0.5}$~\citep{pi05}
and conditions the action expert on the jittered target with the gate $\alpha^{k}$
of the denoising step whose interval $[\tfrac{k-1}{K},\tfrac{k}{K})$ contains $s$.
\begin{algorithm}[h]
\caption{Training step of RACE}
\label{alg:train}
\begin{algorithmic}[1]
\Require observation $\mathbf{o}_t$, instruction $\boldsymbol{\ell}$, ground-truth chunk $\mathbf{A}_t$, soft label $\mathbf{p}_t$
\Ensure training loss $\mathcal{L}$
\State $\mathbf{z}_t \gets \mathrm{VLM}(\mathbf{o}_t,\boldsymbol{\ell})$ \Comment{frozen; cached for both passes}
\State $\boldsymbol{\epsilon} \sim \mathcal{N}(\mathbf{0},\mathbf{I})$
\Statex \textit{Auxiliary pass: unmodulated, at $s=0$}
\State $\mathbf{F}_t \gets$ hidden states of the action expert on $(\boldsymbol{\epsilon},\mathbf{z}_t,s=0)$
\State $\mathbf{v}_t^{\mathrm{aux}} \gets$ velocity predicted from $\mathbf{F}_t$
\State $\hat{\mathbf{p}}_t \gets \mathrm{Head}(\mathbf{F}_t,\hat{\mathbf{z}}_t)$ \Comment{$\hat{\mathbf{z}}_t$: final-layer features in $\mathbf{z}_t$}
\State $\mathcal{L}_{\mathrm{timing}} \gets \mathrm{BCE}(\hat{\mathbf{p}}_t,\mathbf{p}_t)$
\State $\mathcal{L}_{\mathrm{aux}} \gets \|\mathbf{v}_t^{\mathrm{aux}} - (\mathbf{A}_t-\boldsymbol{\epsilon})\|^2$
\Statex \textit{Full pass: modulated, at a sampled flow time}
\State $s \sim p(s)$ \Comment{same distribution as $\pi_{0.5}$}
\State $k \gets \lfloor Ks \rfloor + 1$ \Comment{denoising step whose interval contains $s$}
\State $\mathbf{A}_t^{s} \gets (1-s)\,\boldsymbol{\epsilon} + s\,\mathbf{A}_t$
\State $\tilde{\mathbf{p}}_t \gets \mathrm{Jitter}(\mathbf{p}_t)$ \Comment{teacher forcing; $\pm1$ step w.p.\ $0.5$}
\State $\mathcal{L}_{\mathrm{full}} \gets \|v_{\theta}(\mathbf{A}_t^{s},\mathbf{z}_t,s,\alpha^{k}\tilde{\mathbf{p}}_t) - (\mathbf{A}_t-\boldsymbol{\epsilon})\|^2$
\State \Return $\mathcal{L}_{\mathrm{full}} + \lambda_{\mathrm{aux}}\mathcal{L}_{\mathrm{aux}} + \lambda_{\mathrm{timing}}\mathcal{L}_{\mathrm{timing}}$
\end{algorithmic}
\end{algorithm}

\vspace{-0.15cm}
\paragraph{Inference Procedure.}
At each policy call, the auxiliary pass predicts the prior once,
and the $k$-th of the $K$ denoising steps starts at flow time $s=(k-1)/K$ and uses the gate $\alpha^{k}$.
\begin{algorithm}[h]
\caption{Inference of RACE at policy call $t$}
\label{alg:infer}
\begin{algorithmic}[1]
\Require observation $\mathbf{o}_t$, instruction $\boldsymbol{\ell}$, number of denoising steps $K$
\Ensure action chunk $\mathbf{A}_t$
\State $\mathbf{z}_t \gets \mathrm{VLM}(\mathbf{o}_t,\boldsymbol{\ell})$ \Comment{cached for both passes}
\State $\mathbf{A}_t^{0} \sim \mathcal{N}(\mathbf{0},\mathbf{I})$
\Statex \textit{Auxiliary pass: unmodulated, at $s=0$}
\State $\mathbf{F}_t \gets$ hidden states of the action expert on $(\mathbf{A}_t^{0},\mathbf{z}_t,s=0)$
\State $\hat{\mathbf{p}}_t \gets \mathrm{Head}(\mathbf{F}_t,\hat{\mathbf{z}}_t)$ \Comment{$\hat{\mathbf{z}}_t$: final-layer features in $\mathbf{z}_t$}
\Statex \textit{Full denoising: modulated by the prior}
\For{$k=1,\ldots,K$} \Comment{step $k$ starts at $s=(k-1)/K$}
    \State $\mathbf{A}_t^{k} \gets \mathbf{A}_t^{k-1} + \tfrac{1}{K}\, v_{\theta}\big(\mathbf{A}_t^{k-1},\mathbf{z}_t,\tfrac{k-1}{K},\alpha^{k}\hat{\mathbf{p}}_t\big)$
\EndFor
\State \Return $\mathbf{A}_t^{K}$
\end{algorithmic}
\end{algorithm}

\vspace{-0.1cm}
\section{Further Discussion}
\label{appendix:further_discussion}

\vspace{-0.1cm}
\subsection{Limitations}
\label{app:limitation}
\vspace{-0.1cm}

RACE makes longer action chunks more reliable by learning transition timing
and conditioning action generation on it, but several limitations remain.
\textbf{(1) Imperfect transition timing:}
The prediction head misses some transitions, especially on held-out demonstrations and for longer chunks
(Appendix~\ref{app:accuracy_transition_timing_prediction}),
and its labels are pseudo-labels from change-point detection rather than human annotations.
Nevertheless, even with this imperfect prior, RACE outperforms na\"ive fine-tuning without transition timing at longer chunks
and reduces errors at transitions (Tab.~\ref{tab:vlabench}; Fig.~\ref{fig:drift_error}).
\textbf{(2) Degradation of very long chunks:}
The success rate of RACE declines more steeply beyond $H_{\text{exec}}=20$ and falls below the baseline at $H_{\text{exec}}=40$ (Appendix~\ref{app:scaling_longer_chunks}),
since it reduces errors around transitions but not the error growth of open-loop execution itself.
Still, up to $H_{\text{exec}}=20$, \ie, four times the chunk length of the $\pi_{0.5}$ baseline,
RACE remains comparable to or better than the baseline on all three benchmarks.
\textbf{(3) Reliance on iterative denoising:}
RACE is designed to condition iterative denoising,
and its gain is smaller on a VLA without a denoising process (Appendix~\ref{app:generalization_to_other_vla}).
Even so, it outperforms the fine-tuned variant at every chunk length on both GR00T-N1.6~\citep{gr00t} and OpenVLA-OFT~\citep{openvlaoft}.

\vspace{-0.1cm}
\subsection{Broader Impact}
\label{app:broader_impact}
\vspace{-0.1cm}
This work aims to make robot manipulation with VLA models more continuous
by enabling longer action chunks, which reduces the number of policy calls per task
and the idle time between them.
At the same time, longer chunks mean that the robot executes more actions without new observations,
which gives it fewer chances to react to unexpected changes in its environment,
such as a person entering the workspace.
In addition, RACE is trained on demonstrations,
so it may inherit biases or unsafe behaviors present in the training data.
Our experiments are limited to simulation benchmarks and a controlled laboratory setting.
Deploying such policies in real-world environments, especially near people,
therefore requires additional safeguards, such as safety monitoring, collision checking, and human supervision.

%% file: tab/transition_stats.tex
\begin{table}[h]
\small
\centering
\vspace{-0.2cm}
\caption{Statistics of detected subskill transitions on VLABench training demonstrations.
For each task, we report the number of steps and the number of subskill transitions (\ie, change points) per episode
as mean $\pm$ standard deviation.}
\vspace{-0.15cm}
\label{tab:transition_stats}
\setlength{\tabcolsep}{8pt}
\scalebox{0.97}{
\begin{tabular}{l|c|c|c}
\hline
Task & \# Episodes & Total steps per episode & Transitions per episode \\
\hline
add\_condiment          & 500 & $167.6 \pm 14.1$  & $5.59 \pm 0.80$ \\
insert\_flower          & 500 & $167.3 \pm 15.6$ & $5.42 \pm 0.87$ \\
select\_toy             & 500 & $160.2 \pm 13.7$ & $4.02 \pm 1.09$ \\
select\_chemistry\_tube & 500 & $78.4 \pm 8.6$    & $3.68 \pm 0.69$ \\
select\_drink           & 500 & $97.4 \pm 8.0$    & $3.31 \pm 0.60$ \\
select\_poker           & 500 & $75.4 \pm 7.0$    & $3.06 \pm 0.47$ \\
select\_painting        & 500 & $68.7 \pm 8.9$    & $3.00 \pm 0.56$ \\
select\_book            & 500 & $72.7 \pm 8.6$    & $2.77 \pm 0.66$ \\
select\_mahjong         & 500 & $111.4 \pm 15.4$  & $2.67 \pm 0.88$ \\
select\_fruit           & 500 & $151.3 \pm 15.0$ & $2.19 \pm 1.12$ \\
\hline
All                     & 5,000 & $115.0 \pm 41.8$ & $3.57 \pm 1.35$ \\
\hline
\end{tabular}
}
\vspace{-0.2cm}
\end{table}

%% file: tab/label_validity.tex
\begin{table}[h]
\small
\centering
\vspace{-0.15cm}
\caption{Agreement between transition labels and the reference subskill boundaries
recorded by the VLABench demonstration generator, on 300 newly generated demonstrations with a matching tolerance of $\pm3$ steps.
Precision and recall are counted per transition and pooled over all episodes.
}
\vspace{-0.15cm}
\label{tab:label_validity}
\setlength{\tabcolsep}{8pt}
\begin{tabular}{l|c|cc}
\hline
Transition-label sources & Transitions per episode & Precision & Recall \\
\hline
Random & 5.0 & 0.30 & 0.28 \\
Speed minima~\citep{pace} & 2.4 & 0.40 & 0.21 \\
\hdashline[0.2pt/1pt]
PELT (used) & 3.8 & 0.82 & 0.70 \\
\hline
\end{tabular}
\vspace{-0.15cm}
\end{table}

%% file: tab/libero_comparison.tex
\begin{table}[h]
\small
\caption{Comparison on LIBERO.
$\ast$ PolicyTrim trains a separate model for each suite, and its $H_{\text{exec}}$ is averaged over the four suites. 
Its training cost is also substantially higher; see Appendix~\ref{app:training_efficiency}.
Spd is the per-episode inference speedup relative to the $\pi_{0.5}$ baseline.
$\ddagger$ indicates the average execution length, since adaptive chunking methods truncate each chunk adaptively.
}
\vspace{-0.2cm}
\label{tab:libero_comparison}
\begin{center}
\scalebox{0.92}{
\begin{tabular}{l|@{\hspace{6pt}}c@{\hspace{6pt}}c|cccc|c}
\hline
Method & $H_{\text{exec}}$ & Spd$\uparrow$ & Spatial & Object & Goal & Long & Avg. \\
\hline
$\pi_{0.5}$ {\scriptsize Baseline}
 & 5 & $1.00\times$ & 98.83 & 98.17 & 97.00 & 93.83 & 96.96 \\
\hline
\multicolumn{8}{c}{\textit{Action-Chunk Extension}} \\
\hdashline[0.2pt/1pt]
\multirow{3}{*}{$\pi_{0.5}$ {\scriptsize Fine-tuning}}
 & 10 & $1.98\times$ & 97.60 & 98.80 & 97.60 & 94.40 & 97.10 \\
 & 15 & $2.96\times$ & 97.40 & 99.00 & 97.80 & 93.60 & 96.95 \\
 & 20 & $3.91\times$ & 94.80 & 97.60 & 94.00 & 91.00 & 94.35 \\
    \hdashline[0.2pt/1pt]
    PolicyTrim$^\ast$~\citep{policytrim}
 & 13.75 & $4.85\times$ & 97.80 & 98.50 & 98.80 & 93.30 & 97.10 \\
\hdashline[0.2pt/1pt]
\multirow{3}{*}{RACE (ours)}
 & 10 & $1.95\times$ & 99.40 & 99.60 & 98.60 & 97.00 & 98.65 \\
 & 15 & $2.86\times$ & 99.00 & 99.80 & 98.60 & 96.00 & 98.35 \\
 & 20 & $3.78\times$ & 97.60 &	99.20	& 95.80 &	94.20	& 96.70 \\
\hline
\multicolumn{8}{c}{\textit{Recent State-of-the-Art VLAs}} \\
\hdashline[0.2pt/1pt]
StreamPI~\citep{streampi} & 5 & $0.91\times$ & 98.80 & 99.80 & 99.60 & 95.00 & 98.30 \\
ACoT-VLA~\citep{acotvla} & 5 & $0.68\times$ & 99.40 & 99.60 & 98.80 & 96.00 & 98.45 \\
\hline
\multicolumn{8}{c}{\textit{Adaptive Chunking Methods}} \\
\hdashline[0.2pt/1pt]
AutoHorizon~\citep{autohorizon}
 & 7.10$^\ddagger$ & $1.43\times$ & 97.20 & 99.40 & 98.20 & 94.20 & 97.25 \\
PACE~\citep{pace}
 & 8.96$^\ddagger$ & $1.81\times$ & 98.00 & 98.80 & 97.80 & 94.80 & 97.35 \\
\hline
\hline
\multicolumn{8}{c}{\textit{Other Efficient VLA Methods}} \\
\hdashline[0.2pt/1pt]
SAFE-Pruner~\citep{safepruner}
 & 5 & $1.45\times$ & 97.00 & 98.80 & 96.00 & 90.20 & 95.50 \\
Latent Bridge~\citep{latentbridge}
 & 5 & $1.64\times$ & 99.00 & 97.67 & 97.33 & 93.67 & 96.92 \\
GridS~\citep{grids}
 & 5 & $1.24\times$ & 98.60 & 98.80 & 98.40 & 95.20 & 97.75 \\
    \hline
\end{tabular}
}
\end{center}
\vspace{-0.35cm}
\end{table}

%% file: tab/vlabench_long_chunks.tex
\begin{table}[h!]
\small
\centering
\vspace{-0.15cm}
\caption{
    Results across a wide range of execution chunk lengths on VLABench.
}
\vspace{-0.15cm}
\label{tab:vlabench_long_chunks}
\setlength{\tabcolsep}{5pt}
\resizebox{\textwidth}{!}{%
\begin{tabular}{l|cc|cc|cc|cc|cc|cc|cc}
\hline
\multirow{2}{*}{Method} & \multirow{2}{*}{$H_{\text{exec}}$} & \multirow{2}{*}{Spd$\uparrow$}
& \multicolumn{2}{c|}{In-dist.} & \multicolumn{2}{c|}{Category}
& \multicolumn{2}{c|}{Common.} & \multicolumn{2}{c|}{Instruct.}
& \multicolumn{2}{c|}{Texture} & \multicolumn{2}{c}{Avg.} \\
& & & SR & PS & SR & PS & SR & PS & SR & PS & SR & PS & SR & PS \\
\hline
$\pi_{0.5}$ {\scriptsize Baseline}
    & 5 & $1.00\times$ & 40.4 & 56.7 & 21.4 & 35.6 & 17.0 & 33.7 & 18.0 & 35.9 & 26.0 & 42.1 & 24.6 & 40.8 \\ \hdashline[0.2pt/1pt]
\hline
\multirow{5}{*}{RACE (ours)}
    & 5 & $0.96\times$ & 51.6 & 68.2 & 25.0 & 38.1 & 26.0 & 40.0 & 20.8 & 37.6 & 29.0 & 45.3 & 30.5 & 45.8 \\
    & 10 & $1.92\times$ & 57.6 & 70.9 & 30.4 & 41.8 & 23.2 & 39.0 & 25.0 & 42.0 & 36.0 & 52.8 & 34.4 & 49.3 \\
    & 20 & $3.68\times$ & 54.0 & 68.2 & 27.2 & 38.8 & 28.2 & 41.4 & 21.0 & 39.4 & 34.8 & 51.0 & 33.0 & 47.8 \\
    & 30 & $5.11\times$ & 50.2 & 65.9 & 24.4 & 35.9 & 20.0 & 36.1 & 16.4 & 36.4 & 31.4 & 47.0 & 28.5 & 44.3 \\
    & 40 & $6.82\times$ & 46.0 & 63.1 & 21.8 & 33.3 & 15.4 & 34.0 & 13.0 & 33.0 & 23.8 & 41.0 & 24.0 & 40.9 \\
\hline
\end{tabular}%
}
\vspace{-0.15cm}
\end{table}

%% file: tab/vlm_joint_training.tex
\begin{table}[h]
\small
\centering
\vspace{-0.15cm}
\caption{Effect of training the VLM jointly on VLABench with $H_{\text{exec}}=20$.
VLM frozen indicates our post-training setting, where the VLM is kept frozen and only the action expert is post-trained.
VLM joint training means that the VLM is also updated.
}
\vspace{-0.15cm}
\label{tab:vlm_joint_training}
\setlength{\tabcolsep}{6pt}
\resizebox{\textwidth}{!}{%
\begin{tabular}{l|cc|cc|cc|cc|cc|cc}
\hline
\multirow{2}{*}{Setting}
& \multicolumn{2}{c|}{In-dist.} & \multicolumn{2}{c|}{Category}
& \multicolumn{2}{c|}{Common.} & \multicolumn{2}{c|}{Instruct.}
& \multicolumn{2}{c|}{Texture} & \multicolumn{2}{c}{Avg.} \\
& SR & PS & SR & PS & SR & PS & SR & PS & SR & PS & SR & PS \\
\hline
VLM frozen (used)   & 54.0 & 68.2 & 27.2 & 38.8 & 28.2 & 41.4 & 21.0 & 39.4 & 34.8 & 51.0 & 33.0 & 47.8 \\
VLM joint training         & 57.8 & 72.1 & 30.6 & 43.2 & 28.6 & 43.9 & 31.0 & 50.4 & 46.6 & 62.6 & 38.9 & 54.4 \\
\hline
\end{tabular}
}
\vspace{-0.15cm}
\end{table}

%% file: tab/vlm_joint_training_cost.tex
\begin{table}[h!]
\small
\centering
\vspace{-0.15cm}
\caption{Training cost of the VLM-frozen and VLM joint training settings on VLABench with $H_{\text{exec}}=20$.
Both models are trained for 40K steps on four RTX A6000 GPUs.
Peak memory is the maximum allocated memory per GPU.
}
\vspace{-0.15cm}
\label{tab:vlm_joint_training_cost}
\setlength{\tabcolsep}{4pt}
\begin{tabular}{l|c|c|cc|c}
\hline
Setting & \makecell{Training\\step} & \makecell{Learnable\\params (M)}
& \makecell{Time per\\update (s)} & \makecell{Wall-clock\\time (h)} & \makecell{Peak memory\\(GB/GPU)} \\
\hline
VLM frozen (used) & 40K & 811 & 3.2 & 36 & 18 \\
\hdashline[0.2pt/1pt]
VLM joint training & 40K & 3,734 & 8.6 & 96 & 40 \\
\hline
\end{tabular}
\vspace{-0.15cm}
\end{table}

%% file: tab/pred_exec_length.tex
\begin{table}[h!]
\small
\centering
\vspace{-0.15cm}
\caption{Results with different prediction chunk lengths $H$ at the same execution chunk length $H_{\text{exec}}=20$ on VLABench,
with the $\pi_{0.5}$ baseline as reference.
Only the first $H_{\text{exec}}$ of the $H$ predicted actions are executed per policy call.}
\vspace{-0.15cm}
\label{tab:pred_exec_length}
\setlength{\tabcolsep}{5pt}
\resizebox{\textwidth}{!}{%
\begin{tabular}{l|cc|cc|cc|cc|cc|cc|cc}
\hline
\multirow{2}{*}{Method} & \multirow{2}{*}{$H$} & \multirow{2}{*}{$H_{\text{exec}}$}
& \multicolumn{2}{c|}{In-dist.} & \multicolumn{2}{c|}{Category}
& \multicolumn{2}{c|}{Common.} & \multicolumn{2}{c|}{Instruct.}
& \multicolumn{2}{c|}{Texture} & \multicolumn{2}{c}{Avg.} \\
& & & SR & PS & SR & PS & SR & PS & SR & PS & SR & PS & SR & PS \\
\hline
$\pi_{0.5}$ {\scriptsize Baseline}
    & 10 & 5 & 40.4 & 56.7 & 21.4 & 35.6 & 17.0 & 33.7 & 18.0 & 35.9 & 26.0 & 42.1 & 24.6 & 40.8 \\
\hline
\multirow{2}{*}{$\pi_{0.5}$ {\scriptsize Fine-tuning}}
    & 20 & 20 & 48.8 & 65.3 & 22.8 & 37.2 & 20.2 & 38.2 & 20.8 & 39.3 & 27.8 & 43.9 & 28.1 & 44.8 \\
    & 40 & 20 & 52.6 & 68.2 & 24.0 & 36.5 & 22.4 & 40.5 & 20.8 & 40.9 & 36.0 & 53.7 & 31.2 & 48.0 \\
\hdashline[0.2pt/1pt]
\multirow{2}{*}{RACE (ours)}
    & 20 & 20 & 54.0 & 68.2 & 27.2 & 38.8 & 28.2 & 41.4 & 21.0 & 39.4 & 34.8 & 51.0 & 33.0 & 47.8 \\
    & 40 & 20 & 56.8 & 70.2 & 28.0 & 38.6 & 27.4 & 43.3 & 28.2 & 44.9 & 35.8 & 51.9 & 35.2 & 49.8 \\
\hline
\end{tabular}%
}
\vspace{-0.15cm}
\end{table}

%% file: tab/libero_comparison_other_VLAs.tex
\begin{table}[h]
\centering
\vspace{-0.15cm}
\caption{Generalization to other VLA baselines on LIBERO.}
\vspace{-0.15cm}
\label{tab:backbones}
\begin{subtable}{0.49\textwidth}
\centering
\caption{GR00T-N1.6~\citep{gr00t}}
\vspace{-0.15cm}
\label{tab:backbone_gr00t}
\scriptsize
\setlength{\tabcolsep}{2pt}
\resizebox{0.97\linewidth}{!}{%
\begin{tabular}{l|c|cccc|c}
\hline
Method & $H_{\text{exec}}$ & Spatial & Object & Goal & Long & Avg. \\
\hline
Baseline & 8 & 100.0 & 100.0 & 100.0 & 92.5 & 98.1 \\
\hdashline[0.2pt/1pt]
\multirow{3}{*}{Fine-tuning}
    & 16 & 100.0 & 100.0 &  98.6 & 94.9 & 98.4 \\
    & 24 & 100.0 & 100.0 &  98.4 & 89.6 & 97.0 \\
    & 32 &  98.6 &  98.8 &  95.8 & 90.2 & 95.8 \\
\hdashline[0.2pt/1pt]
\multirow{3}{*}{RACE (ours)}
    & 16 & 100.0 & 100.0 & 100.0 & 97.0 & 99.3 \\
    & 24 & 100.0 & 100.0 & 100.0 & 95.2 & 98.8 \\
    & 32 &  99.0 & 100.0 &  99.0 & 90.0 & 97.0 \\
\hline
\end{tabular}}
\end{subtable}
\hfill
\begin{subtable}{0.49\textwidth}
\centering
\caption{OpenVLA-OFT~\citep{openvlaoft}}
\vspace{-0.15cm}
\label{tab:backbone_openvla}
\scriptsize
\setlength{\tabcolsep}{2pt}
\resizebox{0.97\linewidth}{!}{%
\begin{tabular}{l|c|cccc|c}
\hline
Method & $H_{\text{exec}}$ & Spatial & Object & Goal & Long & Avg. \\
\hline
Baseline & 8 & 97.6 & 98.4 & 97.9 & 94.5 & 97.1 \\
\hdashline[0.2pt/1pt]
\multirow{3}{*}{Fine-tuning}
    & 16 & 96.8 & 99.0 & 96.2 & 92.2 & 96.1 \\
    & 24 & 93.0 & 98.4 & 92.0 & 90.2 & 93.4 \\
    & 32 & 92.4 & 96.4 & 93.6 & 88.6 & 92.8 \\
\hdashline[0.2pt/1pt]
\multirow{3}{*}{RACE (ours)}
    & 16 & 97.4 & 98.5 & 95.8 & 94.4 & 96.5 \\
    & 24 & 93.8 & 97.8 & 93.0 & 92.6 & 94.3 \\
    & 32 & 93.2 & 96.8 & 92.2 & 91.4 & 93.4 \\
\hline
\end{tabular}}
\end{subtable}
\vspace{-0.15cm}
\end{table}

%% file: tab/latency_comparison.tex
\begin{table}[h]
\small
\vspace{-0.15cm}
\caption{Per-call inference latency,
measured on a single NVIDIA RTX A6000 GPU.
$\dagger$ denotes ACoT-VLA components measured in our setup with its official implementation, 
where the explicit action reasoner uses an additional 10-step denoising process to generate coarse action trajectories.
}
\vspace{-0.15cm}
\label{tab:latency_comparison}
\begin{center}
\setlength{\tabcolsep}{8pt}
\begin{tabular}{l|c|c|c}
\hline
Component & $\pi_{0.5}$ & ACoT-VLA & RACE (ours) \\
\hline
VLM                    & 53.5\,ms & 53.5\,ms & 53.5\,ms\\
Action expert (\ie, 10 denoising steps)  & 30.8\,ms & 30.8\,ms & 30.8\,ms \\
\hdashline[0.2pt/1pt]
Explicit action reasoner$^\dagger$ & --   & 32.5\,ms& -- \\
Implicit action reasoner$^\dagger$                 & --   & 3.8\,ms & -- \\ \hdashline[0.2pt/1pt]
Aux.\ one-step denoising pass (Sec.~\ref{sec:head})                & --   & --   & 3.0\,ms\\
Transition-timing head (Sec.~\ref{sec:head})                               & --   & --   & $\sim$0.3\,ms\\
Transition-conditioned modulation (Sec.~\ref{sec:cond})                     & --   & --   & $\sim$0.0\,ms\\
\hline
Total latency                                 & 84.3\,ms & 120.6\,ms & 87.6\,ms \\
Latency ratio to $\pi_{0.5}$                      & $1.00\times$ & $1.43\times$ & $1.04\times$ \\
\hline
\end{tabular}
\vspace{-0.15cm}
\end{center}
\end{table}

%% file: tab/training_efficiency_vlabench.tex
\begin{table}[h]
\small
\centering
\caption{Training efficiency comparison on VLABench.
All methods are measured on four NVIDIA RTX A6000 GPUs, with the compared methods run from their official implementations. 
$\dagger$: following its official setting, ACoT-VLA freezes the LLM backbone and trains the visual encoder,
the action expert, and its components.
}
\vspace{-0.15cm}
\label{tab:training_efficiency_vlabench}
\setlength{\tabcolsep}{3.6pt}
\begin{tabular}{l|cc|c|cc|c}
\hline
Method & \makecell{Training\\step} & \makecell{Global\\batch} & \makecell{Learnable\\params (M)}
& \makecell{Time per\\update (s)} & \makecell{Wall-clock\\time (h)} & \makecell{Peak memory\\(GB/GPU)} \\
\hline
$\pi_{0.5}$ {\scriptsize Fine-tuning} & 40K & 64 & 693 & 2.5 & 28 & 17 \\
\hdashline[0.2pt/1pt]
ACoT-VLA$^\dagger$~\citep{acotvla} & 60K & 128 & 1,309 & 50.6 & 843 & 33 \\
\hdashline[0.2pt/1pt]
RACE (ours) & 40K & 64 & 811 & 3.2 & 36 & 18 \\
\hline
\end{tabular}
\vspace{-0.1cm}
\end{table}

%% file: tab/training_efficiency_libero.tex
\begin{table}[h]
\small
\centering
\vspace{-0.15cm}
\caption{Training efficiency comparison on LIBERO.
All methods are measured on four NVIDIA RTX A6000 GPUs, with PolicyTrim and ACoT-VLA run from their official implementations.
$\dagger$: PolicyTrim trains a separate model for each of the four LIBERO suites with two-stage reinforcement learning
(500 + 500 RL iterations);
its values are for a single suite, and the parentheses give the estimated total for all four suites.
Its time per update is per RL iteration, including rollout collection.
$\ddagger$: following its official setting, ACoT-VLA freezes the LLM backbone and trains the visual encoder,
the action expert, and its components.
}
\vspace{-0.15cm}
\label{tab:training_efficiency_libero}
\setlength{\tabcolsep}{3.3pt}
\begin{tabular}{l|cc|c|cc|c}
\hline
Method & \makecell{Training\\step} & \makecell{Global\\batch} & \makecell{Learnable\\params (M)}
& \makecell{Time per\\update (s)} & \makecell{Wall-clock\\time (h)} & \makecell{Peak memory\\(GB/GPU)} \\
\hline
$\pi_{0.5}$ {\scriptsize Fine-tuning} & 20K & 64 & 693 & 2.5 & 14 & 17 \\
\hdashline[0.2pt/1pt]
PolicyTrim$^\dagger$~\citep{policytrim} & 500 + 500 & 2,048 & 693 & 1,644 & \makecell{457\\{\scriptsize ($\times$4 = 1,828)}} & 36 \\
ACoT-VLA$^\ddagger$~\citep{acotvla} & 20K & 128 & 1,309 & 49.8 & 277 & 33 \\
\hdashline[0.2pt/1pt]
RACE (ours) & 20K & 64 & 811 & 3.0 & 17 & 18 \\
\hline
\end{tabular}
\vspace{-0.1cm}
\end{table}

%% file: tab/seed_variance.tex
\begin{table}[h!]
\small
\centering
\vspace{-0.15cm}
\caption{Statistical significance of RACE on VLABench with $H_{\text{exec}}=20$.
In addition to the model reported in Tab.~\ref{tab:vlabench} (Run 1),
we train RACE twice more with different random seeds (Runs 2 and 3).
The last row reports the mean and standard deviation over the three runs.}
\vspace{-0.15cm}
\label{tab:seed_variance}
\setlength{\tabcolsep}{4pt}
\resizebox{\textwidth}{!}{%
\begin{tabular}{l|c|cc|cc|cc|cc|cc|cc}
\hline
\multirow{2}{*}{RACE} & \multirow{2}{*}{$H_{\text{exec}}$}
& \multicolumn{2}{c|}{In-dist.} & \multicolumn{2}{c|}{Category}
& \multicolumn{2}{c|}{Common.} & \multicolumn{2}{c|}{Instruct.}
& \multicolumn{2}{c|}{Texture} & \multicolumn{2}{c}{Avg.} \\
& & SR & PS & SR & PS & SR & PS & SR & PS & SR & PS & SR & PS \\
\hline
Run 1 {\scriptsize (reported)} & 20 & 54.0 & 68.2 & 27.2 & 38.8 & 28.2 & 41.4 & 21.0 & 39.4 & 34.8 & 51.0 & 33.0 & 47.8 \\
Run 2 & 20 & 53.6 & 68.9 & 28.8 & 40.7 & 29.0 & 45.0 & 22.8 & 40.4 & 35.0 & 52.3 & 33.8 & 49.5 \\
Run 3 & 20 & 52.4 & 66.8 & 32.2 & 42.8 & 25.8 & 40.9 & 23.8 & 43.1 & 30.4 & 49.3 & 32.9 & 48.6 \\
\hdashline[0.2pt/1pt]
Mean $\pm$ std & 20
& 53.3{\scriptsize $\pm$0.8} & 68.0{\scriptsize $\pm$1.1}
& 29.4{\scriptsize $\pm$2.6} & 40.8{\scriptsize $\pm$2.0}
& 27.7{\scriptsize $\pm$1.7} & 42.4{\scriptsize $\pm$2.2}
& 22.5{\scriptsize $\pm$1.4} & 41.0{\scriptsize $\pm$1.9}
& 33.4{\scriptsize $\pm$2.6} & 50.9{\scriptsize $\pm$1.5}
& 33.3{\scriptsize $\pm$0.5} & 48.6{\scriptsize $\pm$0.9} \\
\hline
\end{tabular}}
\vspace{-0.15cm}
\end{table}

%% file: tab/sensitivity_pelt.tex
\begin{table}[H]
\small
\centering
\vspace{-0.15cm}
\caption{Sensitivity to the PELT penalty on VLABench with
$H_{\text{exec}}=20$.
We vary the penalty $\beta$ in Eq.~(\ref{eq:pelt}) of
Appendix~\ref{app:change_point_detection}.}
\vspace{-0.15cm}
\label{tab:sensitivity_pelt}
\setlength{\tabcolsep}{10pt}
\begin{tabular}{c|c|c}
\hline
Penalty $\beta$
& \makecell{\# of transitions\\per episode}
& Avg. SR \\
\hline
2 & 5.10 & 33.1 \\
8 & 1.82 & 32.8 \\
\hdashline[0.2pt/1pt]
4 (used) & 3.57 & 33.0 \\
\hline
\end{tabular}
\vspace{-0.15cm}
\end{table}

%% file: tab/gate_ablation.tex
\begin{table}[H]
\small
\centering
\vspace{-0.15cm}
\caption{Effect of the per-step learnable gate on VLABench with $H_{\text{exec}}=20$.
}
\vspace{-0.15cm}
\label{tab:gate_ablation}
\setlength{\tabcolsep}{10pt}
\begin{tabular}{l|c}
\hline
Gate $\alpha^{k}$ of Eq.~(\ref{eq:race_denoising}) in Sec.~\ref{sec:overview} & Avg. SR\\
\hline
None ($\alpha^{k}=1$) & 30.7 \\
\hdashline[0.2pt/1pt]
Learnable per step (ours) & 33.0 \\
\hline
\end{tabular}
\vspace{-0.15cm}
\end{table}

%% file: tab/soft_label.tex
\begin{table}[H]
\small
\centering
\vspace{-0.15cm}
\caption{Effect of soft transition labels on VLABench with $H_{\text{exec}}=20$.
Hard labels mark only the detected transition step; soft labels spread each transition over neighboring steps with a Gaussian profile (Appendix~\ref{app:soft_label}).
}
\vspace{-0.15cm}
\label{tab:soft_label}
\setlength{\tabcolsep}{8pt}
\begin{tabular}{l|c}
\hline
Transition label & Avg. SR\\
\hline
Hard labels & 32.3 \\
\hdashline[0.2pt/1pt]
Soft labels (used) & 33.0 \\
\hline
\end{tabular}
\vspace{-0.15cm}
\end{table}

%% file: tab/teacher_forcing_jittering.tex
\begin{table}[H]
\small
\centering
\vspace{-0.15cm}
\caption{
Effect of teacher forcing and jittering on VLABench with $H_{\text{exec}}=20$. 
The robustness provided by jittering against inference-time transition shifts is
evaluated separately in Tab.~\ref{tab:sensitivity_shift} of Appendix~\ref{app:sensitivity_to_shift}.
}
\vspace{-0.15cm}
\label{tab:design_choices_training}
\setlength{\tabcolsep}{10pt}
\scalebox{1}{
\begin{tabular}{l|c}
\hline
Training strategy & Avg. SR \\
\hline
w/o teacher forcing & 32.9 \\
w/o jittering & 32.8 \\
\hdashline[0.2pt/1pt]
Both (ours) & 33.0 \\
\hline
\end{tabular}
}
\vspace{-0.15cm}
\end{table}

%% file: tab/timing_accuracy.tex
\begin{table}[H]
\small
\centering
\vspace{-0.15cm}
\caption{Accuracy (\%) of the transition-timing prediction head on VLABench across execution chunk lengths $H_{\text{exec}}$.
\textit{Transition-chunk Detection} evaluates, for each chunk, whether it contains a transition.
\textit{Transition-point Localization} evaluates, for each chunk containing a transition, where the transition occurs.
\textit{Multi-transition Localization} evaluates how many of the transitions within a chunk are captured,
counting every predicted peak rather than only the highest one.
Results are reported on both training and held-out demonstrations.
}
\vspace{-0.15cm}
\label{tab:timing_accuracy}
\setlength{\tabcolsep}{5pt}
\begin{tabular}{ll|ccc|ccc}
\hline
\multirow{3}{*}{Evaluation type} & \multirow{3}{*}{Metric}
& \multicolumn{3}{c|}{Training demos.} & \multicolumn{3}{c}{Held-out demos.} \\
\cline{3-8}
& & \multicolumn{3}{c|}{$H_{\text{exec}}$} & \multicolumn{3}{c}{$H_{\text{exec}}$} \\
& & 10 & 15 & 20 & 10 & 15 & 20 \\
\hline
\multirow{2}{*}{\makecell[l]{Transition-chunk\\Detection}}
  & AUROC & 96.0 & 95.4 & 94.4 & 89.2 & 89.7 & 86.5 \\
  & Acc.  & 87.1 & 87.0 & 86.0 & 79.5 & 79.6 & 77.0 \\
\hdashline[0.2pt/1pt]
\multirow{2}{*}{\makecell[l]{Transition-point\\Localization}}
  & Acc. \scriptsize{within $\pm1$ step} & 80.3 & 75.4 & 72.1 & 68.2 & 58.5 & 50.6 \\
  & Acc. \scriptsize{within $\pm3$ steps} & 90.6 & 85.8 & 83.3 & 81.0 & 69.8 & 65.1 \\
\hdashline[0.2pt/1pt]
\multirow{4}{*}{\makecell[l]{Multi-transition\\Localization}}
  & Precision \scriptsize{within $\pm1$ step} & 86.0 & 86.0 & 87.5 & 84.2 & 81.2 & 80.9 \\
  & Precision \scriptsize{within $\pm3$ steps} & 90.4 & 90.9 & 92.5 & 93.1 & 91.2 & 94.1 \\
  & Recall \scriptsize{within $\pm1$ step} & 67.7 & 65.6 & 62.7 & 54.8 & 52.2 & 50.0 \\
  & Recall \scriptsize{within $\pm3$ steps} & 71.2 & 69.3 & 66.3 & 60.5 & 58.6 & 58.1 \\
\hline
\end{tabular}
\vspace{-0.15cm}
\end{table}

%% file: tab/sensitivity_shift.tex
\begin{table}[H]
\small
\centering
\vspace{-0.15cm}
\caption{Sensitivity to shifts of the predicted transition timing at inference on VLABench with $H_{\text{exec}}=20$.
The peaks of the predicted prior are shifted by $\delta$ steps within the chunk;
peaks shifted beyond the chunk are dropped.
}
\vspace{-0.15cm}
\label{tab:sensitivity_shift}
\setlength{\tabcolsep}{3pt}
\resizebox{\textwidth}{!}{%
\begin{tabular}{l|ccccc|c|ccccc}
\hline
\multirow{2}{*}{Model}
& \multicolumn{11}{c}{Shift $\delta$ of the transition-timing prior} \\
\cline{2-12}
& $-15$ & $-10$ & $-5$ & $-3$ & $-1$ & $0$ & $+1$ & $+3$ & $+5$ & $+10$ & $+15$ \\
\hline
w/o jittering & 28.6 \scriptsize{\textcolor{gray}{$-$4.2}} & 29.8 \scriptsize{\textcolor{gray}{$-$3.0}} & 24.6 \scriptsize{\textcolor{gray}{$-$8.2}} & 24.8 \scriptsize{\textcolor{gray}{$-$8.0}} & 30.1 \scriptsize{\textcolor{gray}{$-$2.7}} & 32.8 & 30.8 \scriptsize{\textcolor{gray}{$-$2.0}} & 26.4 \scriptsize{\textcolor{gray}{$-$6.4}} & 30.4 \scriptsize{\textcolor{gray}{$-$2.4}} & 28.7 \scriptsize{\textcolor{gray}{$-$4.1}} & 31.1 \scriptsize{\textcolor{gray}{$-$1.7}} \\
\hdashline[0.2pt/1pt]
RACE (ours) & 32.1 \scriptsize{\textcolor{gray}{$-$0.9}} & 31.0 \scriptsize{\textcolor{gray}{$-$2.0}} & 29.4 \scriptsize{\textcolor{gray}{$-$3.6}} & 30.3 \scriptsize{\textcolor{gray}{$-$2.7}} & 30.2 \scriptsize{\textcolor{gray}{$-$2.8}} & 33.0 & 30.7 \scriptsize{\textcolor{gray}{$-$2.3}} & 31.2 \scriptsize{\textcolor{gray}{$-$1.8}} & 30.9 \scriptsize{\textcolor{gray}{$-$2.1}} & 30.9 \scriptsize{\textcolor{gray}{$-$2.1}} & 32.1 \scriptsize{\textcolor{gray}{$-$0.9}} \\
\hline
\end{tabular}}
\vspace{-0.15cm}
\end{table}

%% file: tab/prior_intervention_gripper.tex
\begin{table}[H]
\small
\centering
\vspace{-0.15cm}
\caption{Effect of the transition-timing prior on the generated gripper switches on VLABench with $H_{\text{exec}}=20$.
The predicted prior is shifted by $k$ steps or replaced with zeros at inference,
over chunks where the demonstrated gripper command switches exactly once.
Switch offset is the predicted minus the demonstrated switch step.
Values under the predicted prior differ slightly from Tab.~\ref{tab:gripper_timing} due to different noise draws.}
\vspace{-0.15cm}
\label{tab:prior_intervention_gripper}
\setlength{\tabcolsep}{4.5pt}
\begin{tabular}{l|c|ccc|c|ccc}
\hline
\multirow{3}{*}{Prior at inference} & \multicolumn{4}{c|}{Training} & \multicolumn{4}{c}{Held-out} \\
\cline{2-9}
& Switch & \multirow{2}{*}{Exact$\uparrow$} & \multirow{2}{*}{$\pm1$ step$\uparrow$} & \multirow{2}{*}{Missed$\downarrow$}
& Switch & \multirow{2}{*}{Exact$\uparrow$} & \multirow{2}{*}{$\pm1$ step$\uparrow$} & \multirow{2}{*}{Missed$\downarrow$} \\
& offset & & & & offset & & & \\
\hline
Shift $k=-3$ & $-0.64$ & 47.3 & 85.6 & 2.5 & $-0.47$ & 55.0 & 82.2 & 7.5 \\
Shift $k=-2$ & $-0.48$ & 57.9 & 91.3 & 1.1 & $-0.26$ & 60.1 & 89.1 & 5.1 \\
Shift $k=-1$ & $-0.25$ & 71.9 & 94.5 & 1.3 & $+0.06$ & 58.9 & 90.0 & 5.6 \\
\hdashline[0.2pt/1pt]
Predicted (ours) & $-0.05$ & 75.0 & 97.3 & 1.0 & $+0.29$ & 53.8 & 90.8 & 4.1 \\
\hdashline[0.2pt/1pt]
Shift $k=+1$ & $+0.15$ & 72.8 & 95.3 & 2.1 & $+0.50$ & 45.0 & 85.2 & 7.1 \\
Shift $k=+2$ & $+0.49$ & 52.2 & 88.8 & 4.6 & $+0.88$ & 26.3 & 75.9 & 8.8 \\
Shift $k=+3$ & $+0.70$ & 41.0 & 80.3 & 6.4 & $+1.16$ & 16.5 & 62.0 & 10.9 \\
\hline
Zero & $-0.02$ & 62.8 & 85.4 & 8.8 & $+0.36$ & 43.6 & 77.1 & 15.6 \\
\hline
\end{tabular}
\vspace{-0.15cm}
\end{table}

%% file: tab/transition_error_by_component.tex
\begin{table}[H]
\small
\centering
\vspace{-0.15cm}
\caption{Action errors (MSE) at transitions by action component on VLABench with $H_{\text{exec}}=20$.}
\vspace{-0.15cm}
\label{tab:transition_error_by_component}
\setlength{\tabcolsep}{6pt}
\begin{tabular}{l|cc|c}
\hline
\multirow{2}{*}{Error component} & \multicolumn{2}{c|}{MSE} & \multirow{2}{*}{\makecell{Error reduction (\%) over\\$\pi_{0.5}$ {\scriptsize Fine-tuning}}} \\
& $\pi_{0.5}$ {\scriptsize Fine-tuning} & RACE (ours) & \\
\hline
Translation & 3.448 & 2.995 & 13.1 \\
Rotation    & 0.277 & 0.190 & 31.5 \\
Gripper     & 0.124 & 0.094 & 24.6 \\
\hdashline[0.2pt/1pt]
Overall (7-dim) & 1.614 & 1.378 & 14.6 \\
\hline
\end{tabular}
\vspace{-0.15cm}
\end{table}

%% file: tab/gripper_timing.tex
\begin{table}[H]
\small
\centering
\vspace{-0.15cm}
\caption{Gripper switching timing on VLABench with $H_{\text{exec}}=20$,
over chunks where the demonstrated gripper command switches exactly once,
evaluated on 500 randomly sampled training demonstrations and the 300 held-out demonstrations of Appendix~\ref{app:validity_transition_labels}.
All values are in \% except \textit{Mean error}, which is in action steps (0.1\,s each).
\textit{Mean error} is the mean absolute offset between the predicted and the demonstrated switch,
over chunks in which the model predicts a switch; missed switches are counted in the \textit{Missed} column.
\textit{False switch} is the fraction of chunks without a demonstrated switch in which the prediction switches.
\textit{Difference} is RACE minus fine-tuning on the same chunks,
with 95\% percentile confidence intervals of this paired difference in brackets,
obtained from 1,000 bootstrap resamples of episodes.}
\vspace{-0.15cm}
\label{tab:gripper_timing}
\setlength{\tabcolsep}{4pt}
\begin{tabular}{ll|ccc|c|c|c}
\hline
Demos. & Method & Exact$\uparrow$ & $\pm1$ step$\uparrow$ & $\pm3$ steps$\uparrow$ & Missed$\downarrow$ & \makecell{Mean error\\(steps)$\downarrow$} & \makecell{False\\switch$\downarrow$} \\
\hline
\multirow{3}{*}{Training}
& $\pi_{0.5}$ {\scriptsize Fine-tuning} & 70.1 & 94.1 & 97.4 & 2.5 & 0.32 & 0.5 \\
& RACE (ours) & 75.2 & 97.2 & 99.0 & 0.9 & 0.26 & 0.6 \\
& {\scriptsize Difference} & {\scriptsize \makecell{$+5.1$\\{[}3.0, 6.9]}} & {\scriptsize \makecell{$+3.1$\\{[}2.2, 4.0]}} & {\scriptsize \makecell{$+1.6$\\{[}1.0, 2.2]}} & {\scriptsize \makecell{$-1.6$\\{[}$-2.2$, $-1.0$]}} & {\scriptsize \makecell{$-0.06$\\{[}$-0.08$, $-0.04$]}} & {\scriptsize \makecell{$+0.0$\\{[}$-0.1$, 0.2]}} \\
\hline
\multirow{3}{*}{Held-out}
& $\pi_{0.5}$ {\scriptsize Fine-tuning} & 50.6 & 86.9 & 93.7 & 6.3 & 0.54 & 0.6 \\
& RACE (ours) & 53.8 & 91.5 & 95.9 & 3.9 & 0.50 & 0.7 \\
& {\scriptsize Difference} & {\scriptsize \makecell{$+3.2$\\{[}$-1.2$, 7.6]}} & {\scriptsize \makecell{$+4.6$\\{[}1.9, 7.1]}} & {\scriptsize \makecell{$+2.2$\\{[}0.5, 4.1]}} & {\scriptsize \makecell{$-2.4$\\{[}$-4.3$, $-0.9$]}} & {\scriptsize \makecell{$-0.04$\\{[}$-0.09$, 0.01]}} & {\scriptsize \makecell{$+0.1$\\{[}$-0.1$, 0.3]}} \\
\hline
\end{tabular}
\vspace{-0.15cm}
\end{table}

%% file: tab/idle_decomposition.tex
\begin{table}[h]
\small
\centering
\vspace{-0.15cm}
\caption{Breakdown of the idle time in Tab.~\ref{tab:realworld_pnp} over successful trials.
The robot keeps executing the current chunk while the next one is prepared,
and pauses only when the latency exceeds this execution buffer,
\ie, the execution time of the actions remaining in a chunk when it arrives,
after real-time chunking skips the actions elapsed during inference.
Latencies are averaged over calls and, unlike Tab.~\ref{tab:latency_comparison} for latency overhead comparisons,
include camera capture, network transfer, and real-time chunking~\citep{rtc}.
Pause per call exceeds Latency $-$ buffer because calls within the buffer count as zero pause.}
\vspace{-0.15cm}
\label{tab:idle_decomposition}
\setlength{\tabcolsep}{6pt}
\begin{tabular}{l|cc|c}
\hline
& \multicolumn{2}{c|}{$\pi_{0.5}$} & RACE (ours) \\
& $H_{\text{exec}}=5$ & $H_{\text{exec}}=20$ & $H_{\text{exec}}=20$ \\
\hline
\# of policy calls per episode & 45 & 17 & 17 \\
Actions executed per call & 7.0 & 19.9 & 20.1 \\
\hdashline[0.2pt/1pt]
\textbf{Latency per call} & \textbf{271\,ms} & \textbf{285\,ms} & \textbf{372\,ms} \\
\quad Camera capture & 122\,ms & 112\,ms & 109\,ms \\
\quad Network & 49\,ms & 64\,ms & 75\,ms \\
\quad Server & 100\,ms & 109\,ms & 188\,ms \\
\textbf{Execution buffer} & \textbf{234\,ms} & \textbf{1160\,ms} & \textbf{1072\,ms} \\
\quad Actions skipped on arrival & 3.0 & 5.3 & 7.8 \\
\hdashline[0.2pt/1pt]
Latency $-$ buffer & $+$37\,ms & $-$875\,ms & $-$700\,ms \\
Pause per call & 39\,ms & 0\,ms & 0\,ms \\
\textbf{Idle time} & \textbf{2.11\,s} & \textbf{0.40\,s} & \textbf{0.41\,s} \\
\quad Startup (first camera capture and call) & 0.27\,s & 0.28\,s & 0.28\,s \\
\quad Pauses after the first call & 1.71\,s & 0.00\,s & 0.00\,s \\
\quad Control loop & 0.13\,s & 0.12\,s & 0.13\,s \\
\hline
\end{tabular}
\vspace{-0.15cm}
\end{table}

%% file: tab/speedup_breakdown.tex
\begin{table}[h]
\small
\centering
\vspace{-0.15cm}
\caption{Breakdown of the inference speedup at $H_{\text{exec}}=20$ on VLABench,
over the 127 episodes that all three models complete successfully.
Spd is computed over all episodes pooled and with each task weighted equally;
the last column shows its range across tasks.}
\vspace{-0.15cm}
\label{tab:speedup_breakdown}
\setlength{\tabcolsep}{4pt}
\begin{tabular}{l|cc|ccc}
\hline
\multirow{2}{*}{Model} & \multirow{2}{*}{\makecell{Env. steps\\per episode}} & \multirow{2}{*}{\makecell{Policy calls\\per episode}}
& \multicolumn{3}{c}{Spd$\uparrow$} \\
\cline{4-6}
& & & Pooled & Task-balanced & Range across tasks \\
\hline
$\pi_{0.5}$ {\scriptsize Baseline} & 117.9 & 23.93 & $1.00\times$ & $1.00\times$ & -- \\
\hdashline[0.2pt/1pt]
$\pi_{0.5}$ {\scriptsize Fine-tuning} & 112.3 & 6.10 & $3.92\times$ & $3.89\times$ & $3.15$--$4.70\times$ \\
RACE (ours) & 113.6 & 6.16 & $3.74\times$ & $3.64\times$ & $2.54$--$4.53\times$ \\
\hline
\end{tabular}
\vspace{-0.15cm}
\end{table}

%% file: tab/speedup_sets.tex
\begin{table}[h]
\small
\centering
\vspace{-0.15cm}
\caption{Inference speedup at $H_{\text{exec}}=20$ on VLABench under different sets of episodes.
The first row is the setting used in the main comparisons.}
\vspace{-0.15cm}
\label{tab:speedup_sets}
\setlength{\tabcolsep}{6pt}
\begin{tabular}{l|c|cc}
\hline
Episode set & \# Episodes & $\pi_{0.5}$ {\scriptsize Fine-tuning} & RACE (ours) \\
\hline
Common successes with the $\pi_{0.5}$ baseline (used) & 228 / 256 & $3.82\times$ & $3.68\times$ \\
\hdashline[0.2pt/1pt]
Common successes of all three models & 127 & $3.92\times$ & $3.74\times$ \\
\quad with each task weighted equally & 127 & $3.89\times$ & $3.64\times$ \\
All episodes, regardless of success & 2,500 & $4.00\times$ & $3.94\times$ \\
\hline
\end{tabular}
\vspace{-0.15cm}
\end{table}

%% file: tab/matched_action_change.tex
\begin{table}[H]
\small
\centering
\vspace{-0.15cm}
\caption{Ratio of the action error at transitions to that at non-transition positions
matched by position within the chunk and by decile of the second-order action difference $c_h$,
on VLABench with $H_{\text{exec}}=20$.
A ratio of 1 means no extra error at transitions.
Brackets denote 95\% episode-level bootstrap confidence intervals.}
\vspace{-0.15cm}
\label{tab:matched_action_change}
\setlength{\tabcolsep}{6pt}
\begin{tabular}{l|cc}
\hline
& $\pi_{0.5}$ & RACE (ours) \\
\hline
Error ratio (transition / matched)
& 1.34 {\scriptsize [1.15, 1.56]} & 1.30 {\scriptsize [1.05, 1.62]} \\
\hline
\end{tabular}
\vspace{-0.15cm}
\end{table}